\documentclass[manuscript,screen]{acmart-preprint}
\usepackage{subcaption}
\usepackage{adjustbox}

\AtBeginDocument{%
  \providecommand\BibTeX{{%
    \normalfont B\kern-0.5em{\scshape i\kern-0.25em b}\kern-0.8em\TeX}}}

\setcopyright{none}

\begin{document}

\title{Multimodal Classification: Current Landscape, Taxonomy and Future Directions}

\author{William C. Sleeman IV}
\email{wcsleeman@vcu.edu}
\orcid{0000-0002-8431-8483}
\affiliation{%
  \institution{Virginia Commonwealth University}
  \streetaddress{601 West Main Street}
  \city{Richmond}
  \state{Virginia}
  \country{USA}
  \postcode{23284-3068}
}

\author{Rishabh Kapoor}
\email{rishabh.kapoor@vcuhealth.org}
\affiliation{%
  \institution{Virginia Commonwealth University}
  \streetaddress{601 West Main Street}
  \city{Richmond}
  \state{Virginia}
  \country{USA}
  \postcode{23284-3068}
}

\author{Preetam Ghosh}
\email{pghosh@vcu.edu}
\affiliation{%
  \institution{Virginia Commonwealth University}
  \streetaddress{601 West Main Street}
  \city{Richmond}
  \state{Virginia}
  \country{USA}
  \postcode{23284-3068}
}

\renewcommand{\shortauthors}{Sleeman et al.}

\begin{abstract} Multimodal classification research has been gaining popularity in many domains that collect more data from multiple sources including satellite imagery, biometrics, and medicine. However, the lack of consistent terminology and architectural descriptions makes it difficult to compare different existing solutions. We address these challenges by proposing a new taxonomy for describing such systems based on trends found in recent publications on multimodal classification. Many of the most difficult aspects of unimodal classification have not yet been fully addressed for multimodal datasets including big data, class imbalance, and instance level difficulty. We also provide a discussion of these challenges and future directions.

\end{abstract}




\maketitle

\section{Introduction}
	\label{sec:int}	
	In the last decade, there has been an increased focus on combining data from multiple modalities to further improve machine learning based classification models. By using information from several representations of the same subject, a more complete picture of the problem at hand can be constructed. Multiple data modalities are naturally present in many problem domains such as medicine \cite{jiang2021review,huang2020review,haskins2020deep,garcia2018mental}, hyperspacial imagery \cite{audebert2019deep, gu2017multiple}, sentiment analysis \cite{soleymani2017survey,chandrasekaran2021multimodal,sharma2020multimodal}, and many others. However, combining diverse data modalities can be challenging when each individual representation is significantly different and can include data combinations of image-text, audio-video, or multiple sensors that are not time synchronized.
	
	The increasing interest in multimodal learning has led to a number of recent survey papers covering entire domains \cite{baltruvsaitis2018multimodal, zhao2017multi, li2018survey,zhang2020multimodal}. However, there has been little work on the inherent challenges facing multimodal classification problems as many of the existing surveys focused specifically on deep learning \cite{ramachandram2017deep, yan2021deep, guo2019deep}, domain specific solutions \cite{audebert2019deep, garcia2018mental, griffiths2019review}, or non-classification methods \cite{raman2018multimodal, jiang2021review, haskins2020deep}. 
		
	In this survey, we provide a review of recent works in multimodal classification and observations on the architectures most commonly used. Unlike other works, we also investigate multimodal classification applications that include both traditional machine learning and deep learning models. Problems related to distributed computing, class imbalance, and instance level difficulty have been well researched for unimodal learning but have not been addressed in the multimodal context. We provide an in-depth discussion of these challenges, the specific issues they raise with multimodal learning, and explore areas prime for future research.
	
	Since there are many different ways to describe multimodal classification models, inconsistencies in their terms and overall presentations make it difficult to easily compare different approaches. Using the models from these reviewed papers, we also present a holistic taxonomy for describing the general architecture of multimodal classification models and their primary components. By distilling the common architectural design patterns found in current multimodal classification research, we provide a high level scheme for describing such systems. In summary, the main contributions of this survey includes:
	\label{sec:contributions}
		\begin{itemize}
		\item A review of recent multimodal classification research
		\item An overview of commonly used multimodal architectures, both with traditional machine learning and deep learning methods
		\item A proposed holistic taxonomy for multimodal classification architectures
		\item A discussion of the open challenges with multimodal classification
	\end{itemize}

	In Section \ref{sec:mvl}, we provide a brief overview of multimodal learning and examples of existing domain specific solutions. Section \ref{sec:arch} gives descriptions of multimodal classification architectures and our proposed taxonomy, while Section \ref{sec:classification} reviews recent multimodal classification research using these common terms. Section \ref{sec:limits} discusses challenges with classification that have not been addressed for multimodal problems and Section \ref{sec:con} provides our concluding remarks. In Table ~\ref{table:terms}, we introduce the list of abbreviations that will be frequently used in the rest of this paper.
	
		\begin{table}[ht]
		\centering
		\renewcommand{\arraystretch}{1.25}
		\begin{adjustbox}{center,max width = 135mm}
			\footnotesize
			\begin{tabular}{ll}\hline
				\multicolumn{2}{c}{\textbf{List of Terms}}\\
				\bottomrule
				\textbf{Term} & \textbf{Description} \\
				\bottomrule
				AE & Autoencoder\\
				CNN & Convolutional neural network\\
				DBN & Deep believe network\\			
				DNN & Deep neural network\\
				DT & Decision tree\\
				ELM & Extreme learning machine\\
				GBT & Gradient boosted trees\\
				GRU & Gated recurrent unit\\
				KNN & \textit{k}-nearest neighbors\\
				LR & Linear regression\\		
				LSTM & Long short-term memory\\											
				Manual & Feature extraction performed by hand or with a manually chosen method\\
				MKL & Multiple kernel learning\\
				MLP & Multilayer perceptron\\
				Multi-SVNN & Multi support vector neural network\\
				RF & Random forest\\
				SVD & Singular value decomposition\\
				SVM & Support vector machine\\
				RNN & Recurrent neural network\\
				WE & Word embedding\\
				XGBoost & Extreme gradient boosting\\								
				\bottomrule
			\end{tabular}
		\end{adjustbox}
		\caption{A list of terms used to describe stages in multimodal classification architectures.}
		\label{table:terms}
	\end{table}

	\section{Previous Multimodal Research}
	\label{sec:mvl}
	Prior mulimodal surveys and reviews have covered a collection of domain specific problems, fusion approaches, network architectures, and different kinds of learning other than classification. In this section, we review the core concepts in multimodal learning, existing taxonomies, and domain specific research.

	\subsection{Representations and Tasks}
	\label{sec:representations}
	According to the taxonomy defined by \cite{baltruvsaitis2018multimodal}, multimodal learning can be broken down into the following five challenges: representation \cite{baltruvsaitis2018multimodal, zhang2020multimodal}, translation \cite{baltruvsaitis2018multimodal}, alignment \cite{baltruvsaitis2018multimodal,li2018survey, sun2013survey}, fusion \cite{baltruvsaitis2018multimodal,sun2013survey, li2018survey,zhang2020multimodal}, and co-learning \cite{baltruvsaitis2018multimodal, zhao2017multi}.\newline
	
	\noindent \textbf{Representation} challenges are related to the heterogeneity between modalities. For example, combining text and image data may be non-trivial as their formats are significantly different.

	\noindent \textbf{Translation} is used to map from one modality to another. Text descriptions can be generated from image data but the process is complicated by one-to-many scenarios as there are likely multiple valid mappings from a single image.
	 
	\noindent \textbf{Alignment} occurs when two or more modalities need to be synchronized, such as matching audio to video frames.
	 
	\noindent \textbf{Fusion} is the method for combining data from multiple modalities before applying a learning algorithm.

	\noindent \textbf{Co-learning} is a form of transfer learning where data from one modality can be used to boost the other. This can include the generation of synthetic examples to correct missing data or to help the learning of one modality with shared information.\newline
	
	In addition to supervised classification, which is addressed in the remaining sections, a number of other multimodal learning tasks have been researched. Prior surveys have discussed semi-supervised learning \cite{zhao2017multi, xu2013survey, sun2013survey}, encoding \cite{gao2020survey}, clustering \cite{zhao2017multi, sun2013survey}, and multi-task learning \cite{zhao2017multi, li2018hierarchical}.

	\subsection{Taxonomies}
	One of the earlier taxonomies \cite{sun2013survey} divided multimodal learning into co-training and co-regularization methods. Later, \cite{baltruvsaitis2018multimodal} provided a more comprehensive taxonomy as discussed in the Section \ref{sec:representations}. Data fusion is a core concept of all multimodal approaches and \cite{baltruvsaitis2018multimodal} also grouped them as model-agnostic (early, late, and hybrid fusion) and model-based (kernel-based, graphical models, neural networks) methods. The early, late, and hybrid fusion methods were also discussed by \cite{di2018signals} and \cite{simonetta2019multimodal}. In addition to general taxonomies, several were defined for specific domains such as image matching \cite{jiang2021review}, mental health monitoring systems (MHMS) \cite{garcia2018mental}, and human activity research \cite{di2018signals}.
	
	\subsection{Domain Specific Solutions}
	One of the common applications for mutlimodal learning is remote sensing with hyperspectral satellite imagery. This method collects image data from the target area using multiple light wavelengths such as standard RGB and infrared or can include imaging technologies like LiDAR. In one review paper \cite{gu2017multiple}, multiple kernel learning approaches were investigated for image classification. These kernel methods map input data to a new feature space which then can be used by SVM-based classifiers, resulting in something similar to late fusion architectures as discussed in Section \ref{sec:late}. Focusing on deep learning methods, \cite{audebert2019deep} covered different networks designed for hyperspectral classification. The authors observed that 2-D approaches work well on data with spatial relationships and 3-D approaches for hyperspectral data where the third dimension represents image modalities. It was also suggested that Gaussian mixture models and GANs can be used to augment training data by approximating the embedded space. Results of the 2017 IEEE Geoscience and Remote Sensing Society Data Fusion Contest showed that the top teams all utilized data from multiple sources and used ensemble methods \cite{yokoya2018open}.

	In a similar fashion, multimodal learning has also been applied to medical imaging. Today, it is common for multiple image modalities like computed tomography (CT), magnetic resonance imaging (MRI), or positron emission tomography (PET) to be fused together to provide additional information for determining a diagnosis or the best treatment procedure. A number of image fusion architectures were reviewed by \cite{huang2020review} and they observed a current limitation that few existing fusion methods utilized more than two image modalities at the same time. A survey by \cite{haskins2020deep} also covered medical image fusion while comparing both rigid and deformable registration techniques. To address unrealistic image deformations used for non-linear image registrations, GANs were proposed as they often can learn to generate plausible synthetic images. In both papers, it was mentioned that the lack of standard evaluation metrics makes accurate assessment of image fusion methods difficult. The field of neuroimaging has also utilized multimodal imagery to improve scientific understanding and diagnostic performance \cite{dahne2015multivariate}, including two surveys by \cite{rathore2017review} and \cite{liu2018use} that focused on Alzheimer’s disease.
	
	Another problem space well adapted for multimodal learning is human activity tracking. With the reduced cost and size, wearable sensors including microphones, accelerometers, and GPS are now practical to use. In one work \cite{radu2018multimodal}, deep learning techniques for activity and context recognition were investigated. Several neural network architectures with the traditional early and late fusion methods were evaluated as well as different feature extraction and modality data combiners. The authors also mentioned a challenge with features extraction as it was not clear if signal data should be treated as time domain points or be further processed with methods like a fast Fourier transform (FFT). In another survey \cite{garcia2018mental}, wearable sensors were used to monitor mental health conditions with traditional machine learning algorithms. Based on modalities used in prior driver stress detection systems, \cite{rastgoo2018critical} proposed a multimodal framework using various types of sensors.
		
	Other domain specific surveys have also covered biometrics, 3-D image classification, and music information processing. The field of biometrics uses human features, such as images of the face, ear, or fingerprints, to identify individuals and \cite{oloyede2016unimodal} reviewed different multimodal architectures and fusion methods for that domain.  \cite{griffiths2019review} also reviewed works on 3-D object classification using multimodal inputs. By using different representations of objects, such as with 2-D images taken from multiple angles or RGB plus depth (RGB-D) images, multimodal models were used for identification. In addition to the presentation of many different network architectures, it was observed that multimodal 2-D models can perform well on a 3-D task, especially since pre-trained 2-D networks were more mature than 3-D networks. \cite{zhang2020deep} also performed a review of research using multimodal image data such as RGB-D for image segmentation. In the context of music processing, \cite{simonetta2019multimodal} explored preprocessing steps like modality synchronization, feature extraction methods, and the conversion of multiple modalities to a common feature space.
	
	\section{Multimodal Classification Taxonomy} 
	\label{sec:arch}
	
	A current challenge with multimodal learning research is the wide mix of terms describing different aspects of the learning process. Many of the previously discussed papers used terms like \textit{early}, \textit{late}, \textit{intermediate}, or \textit{hybrid} to describe such architectures but their definitions are not always the same. Today, practitioners of deep learning methods are immediately familiar with networks described in terms of CNNs, GANs, or fully connected architectures but those kinds of portrayals are not present for multimodal classification architectures. In this section, we propose a new multimodal classification taxonomy to provide high-level, but descriptive, terms that can be used to more easily compare such models.
	
	From the reviewed works in Tables \ref{table:machinelearning} and \ref{table:deeplearning} and the machine learning \cite{xu2013survey} and deep learning \cite{ramachandram2017deep, gao2020survey, guo2019deep, yan2021deep, zhang2020multimodal} surveys, we propose a taxonomy with five major stages used for building multimodal classification models: preprocessing, feature extraction, data fusion, primary learning, and classification. In some cases, multiple stages are essentially combined into a single step of a model but all of the underlying operations are still performed. In Section \ref{sec:classification}, we will present recent research using these terms, discuss some of scenarios where multiple architectural concepts are employed in a single multimodal architecture, and where the exact architectural description is more subjective.


	

	
	

	\subsection{Preprocessing}
	\label{sec:preprocessing}
	Although not always used, many classification models require some preprocessing, whether it be addressing missing data values, cropping images, or filtering noise. Here, we describe preprocessing as a data cleaning step, done with some level of domain expertise which may be difficult to generalize in this proposed taxonomy. While there are many ways to clean unimodal data, even more options are possible with multimodal datasets if each modality is processed independently. For example, if CT and MRI data is used, different strategies for cropping, scaling, and noise reduction may be required. These images may also need to be registered, or aligned, using a rigid or deformable transformation. However, all preprocessing could also be skipped for one or both modalities to use the raw data instead. Because this work is very user and domain dependent, the preprocessing step is not further discussed in detail but should be considered in practice at a case-by-base basis. 

	\subsection{Feature Selection}
	\label{sec:selection}
	Each multimodal classification model uses feature selection in some capacity and may include manual feature engineering, deep learning methods, or be an inherent part of a classifier algorithm. The feature selection process can be performed independently for each modality or be part of multiple steps in the overall model architecture. Deep learning methods like CNNs are often used for feature extraction but the same network may also perform the classification step, thus doing two tasks at once. Although classifiers like Random Forests can perform the feature selection process by identifying the most useful cut points during the creation of decision trees, this operation could be performed at one explicit step, such as with a CNN for image feature extraction followed by a DNN classifier. Dimensionality reduction using Principal Component Analysis (PCA) or Linear Discriminant Analysis (LDA) can be used as part of the features selection process, most commonly used in traditional machine learning models which can struggle with very high dimensional data.

	\subsection{Data Fusion}
	\label{sec:fusion}		
	Data fusion is a unique aspect of multimodal learning where information from different data sources need to be combined when building a joint model. This process may happen right after the input data is presented, right before the final classification, or multiple times in the middle. Commonly used terms for these architectures include early or late fusion but these terms may not be enough to fully describe a multimodal model. Since previous works presented these fusion approaches differently, we propose a series of definitions which will be used throughout this paper.

	\subsubsection{Early Fusion}
	\label{sec:early}
\begin{figure}[ht]
		\centering
		\includegraphics[height=4cm]{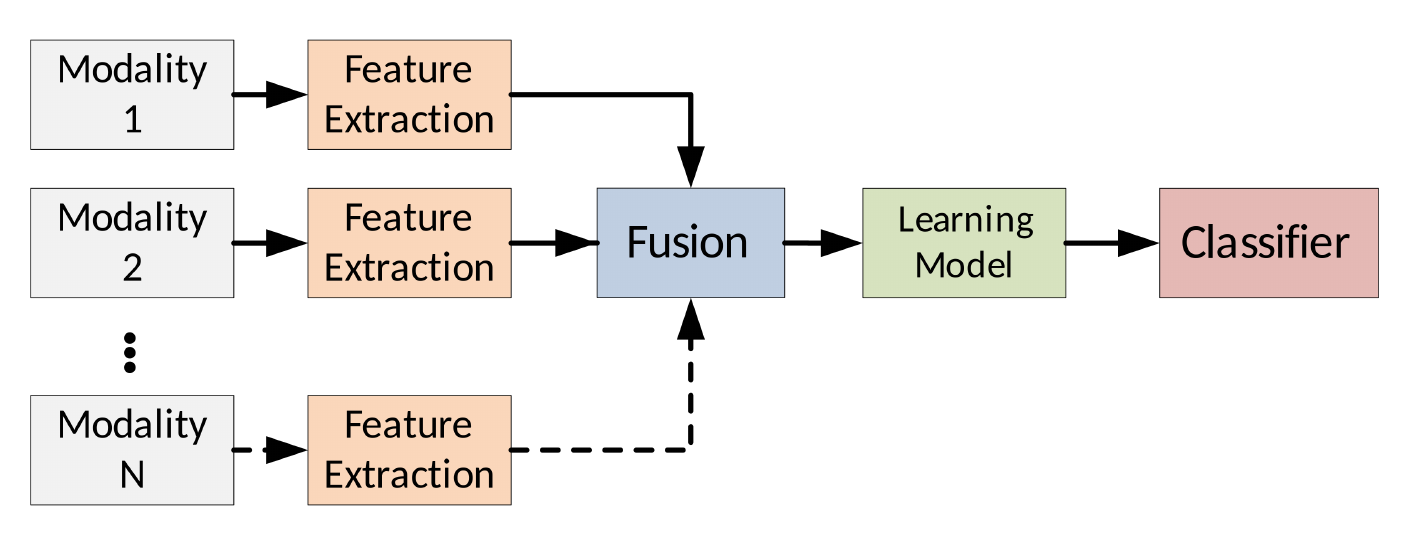}
		\caption{An example architecture for multimodal classification with early fusion.}
		\label{fig:early}
	\end{figure}

	Early fusion occurs when all of the multimodal data is merged before the primary learning model has been performed. As shown in Figure \ref{fig:early}, data from two modalities is joined and then passed to a classifier. One of the most common ways to achieve this kind of fusion is to simply concatenate the incoming modality data which can include traditional feature vectors or output nodes from pre-training neural networks. Each modality could also represent a different channel in a CNN model and may be most appropriate when there is a strong association between each data source. For example, radiotherapy datasets with imaging (CT) and planning dose volumes can be stacked as CNN channels since each modality usually has a one-to-one voxel (3-D pixel) relationship. Satellite imagery using different light wavelengths could also be fused in a similar manner if each modality is representing the same ground area.
	
	\subsubsection{Late Fusion}
	\label{sec:late}
\begin{figure}[ht]
		\centering
		\includegraphics[height=4cm]{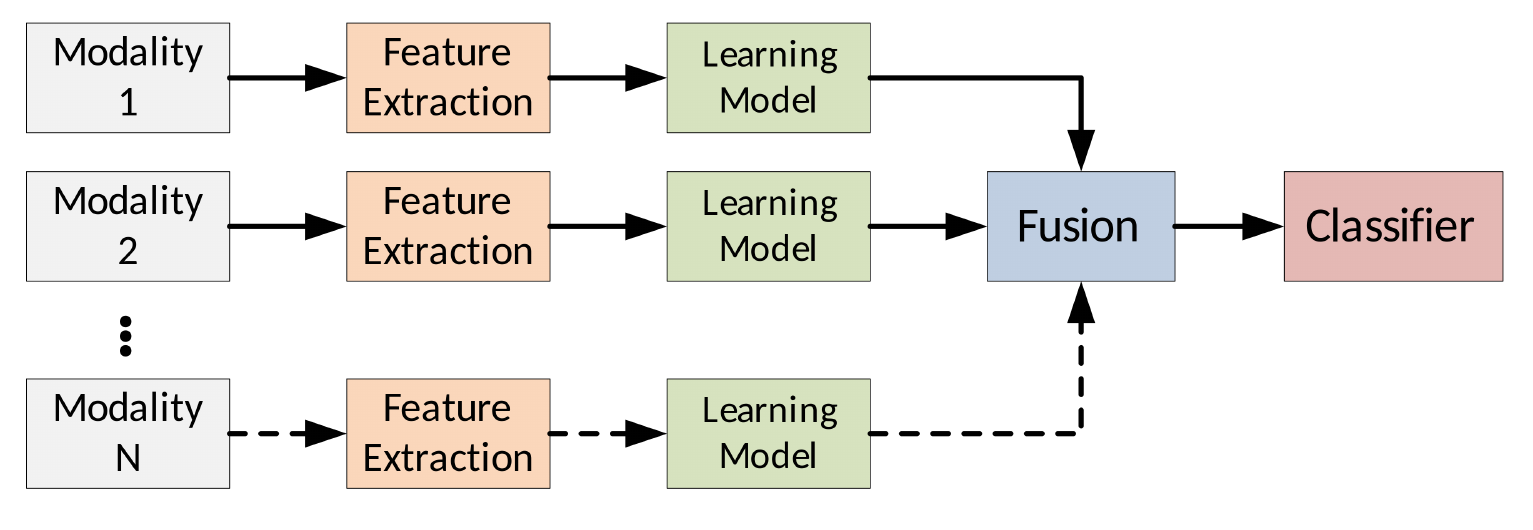}
		\caption{An example architecture for multimodal classification with late fusion.}
		\label{fig:late}
	\end{figure}

	With late fusion, feature extraction and the modality specific learning is performed independently before the final classification as shown in Figure \ref{fig:late}. Output from the fusion stage can include low level learned features with deep networks or class probabilities from full classifier algorithms. In both cases, the learned results are combined for the final classification. This architecture benefits from the ability to train each modality with a specific algorithm and may make it easier to add or exchange different modalities in the future. One downside is the lack of cross-modality data sharing which could hinder learning the relationships between modalities.

	\subsubsection{Cross-modality Fusion}
	\label{sec:intermediate}
\begin{figure}[ht]
		\centering
 		\includegraphics[height=4cm]{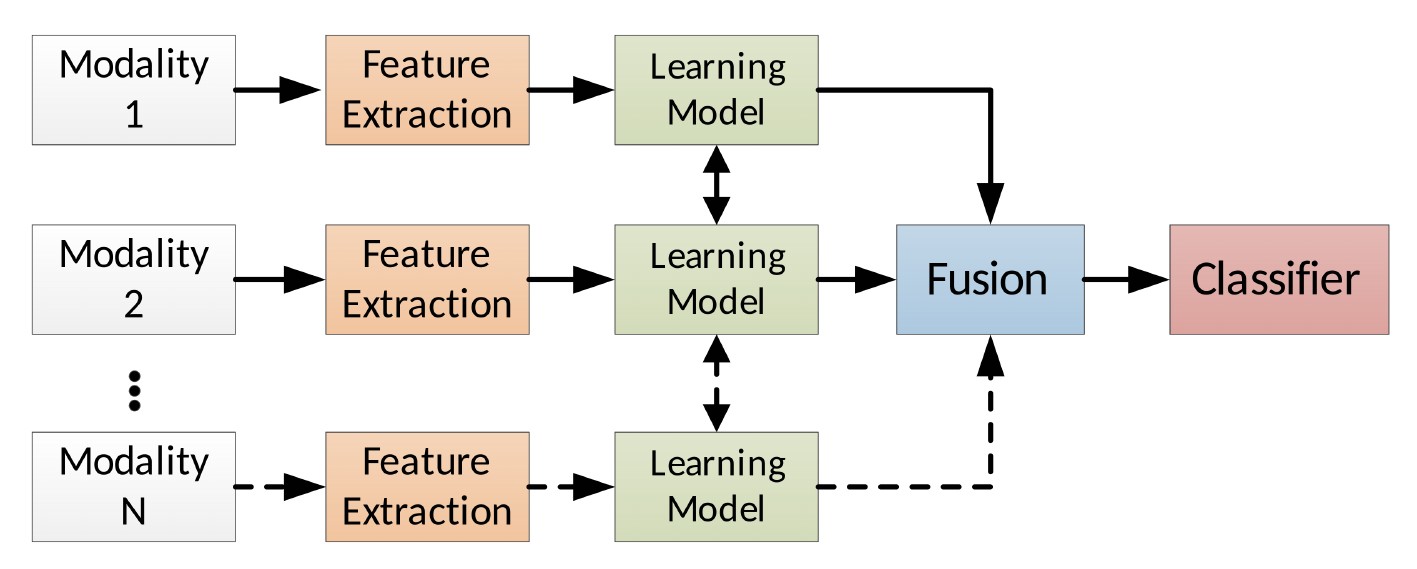}
		\caption{An example cross-modality architecture with a single data sharing operation.}
		\label{fig:cross1}
	\end{figure}

\begin{figure}[ht]
		\centering
		\includegraphics[height=3cm]{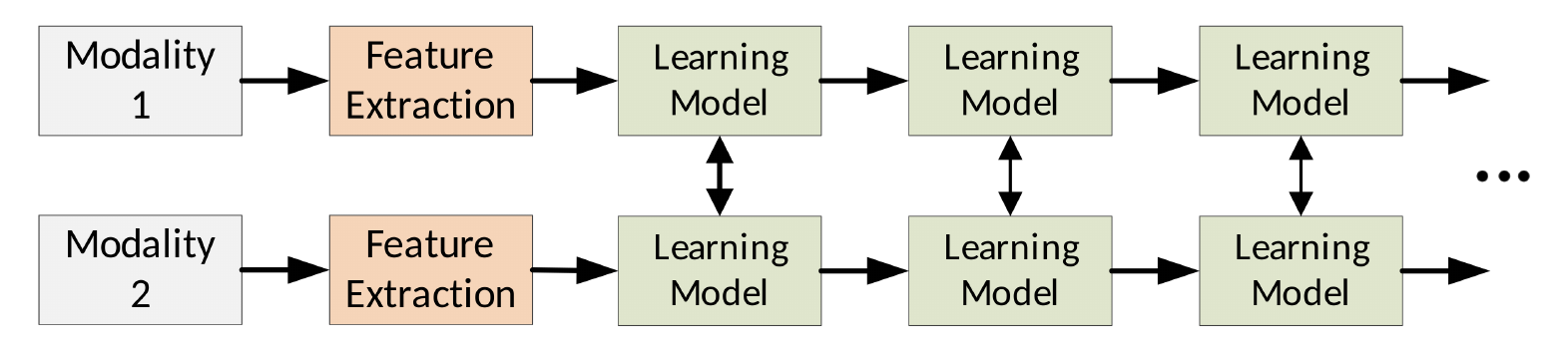}
		\caption{The first part of a sample cross-modality architecture where data between modalities is shared multiple times during the learning process.}
		\label{fig:cross2}
	\end{figure}

    Cross-modality fusion allows for the sharing of modality specific data before or during the primary learning stage. Unlike early or late fusion, this approach provides a way that each modality can use the context of each other to improve the predictive power of the overall model. This data sharing can be represented in many ways including at different parts of the learning process, the type or amount of data shared, or which modalities participate in the sharing. Figure \ref{fig:cross1} shows an architecture sharing data between each modality once before fusion and Figure \ref{fig:cross2} shows data sharing occurring multiple times during training. A number of the presented papers showed that this kind of data sharing can out perform the traditional early or late fusion approaches, suggesting a promising direction for solving multimodal problems. In one work classifying satellite images \cite{hong}, each modality was partially trained using a CNN and the results were merged with the original data from the other modality. Another set of CNNs were used to continue the learning of those combined features and results were merged again before performing the final classification. Similar to the sharing style shown in Figure \ref{fig:cross2}, the model presented by \cite{gao} shared partially learned feature between parallel CNN networks multiple times for Alzheimer’s disease classification. The results from the last stage of each modality specific network was concatenated before making predictions. This cross-modality fusion architecture is often also used with deep belief network (DBN) or autoencoder style networks as depicted by \cite{gao2020survey, guo2019deep, li2018survey}.

	\subsection{Data Joining Styles}
	\label{sec:mixing}
	
\begin{figure}[ht]
		\centering
		\begin{subfigure}{.5\textwidth}
			\centering
			\includegraphics[width=0.75\textwidth]{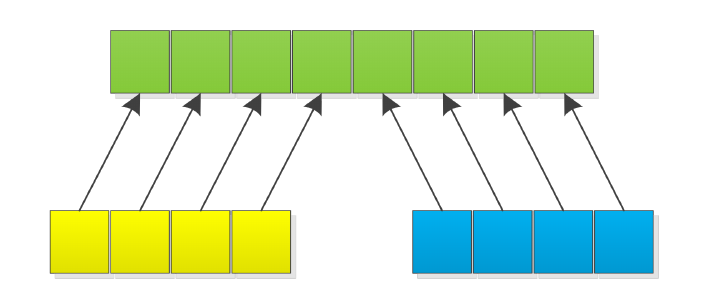}
			\caption{Feature Concatenation}
			\label{fig:feature_concat}
		\end{subfigure}%
		\begin{subfigure}{.5\textwidth}
			\centering
			\includegraphics[width=0.75\textwidth]{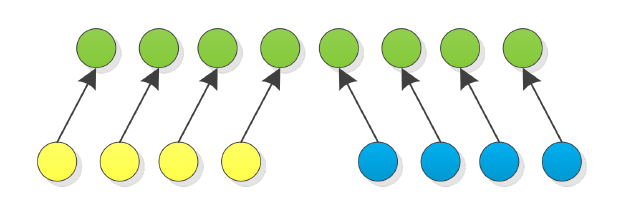}
			\caption{Deep Concatenation}
			\label{fig:deep_concat}
		\end{subfigure}%
	
		\begin{subfigure}{.5\textwidth}
			\centering
			\includegraphics[width=0.75\textwidth]{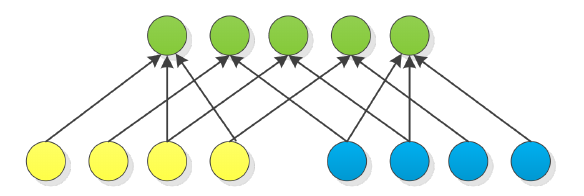}
			\caption{Deep Merge}
			\label{fig:deep_merge}			
		\end{subfigure}
		\begin{subfigure}{.4\textwidth}
			\centering
			\includegraphics[width=0.75\textwidth]{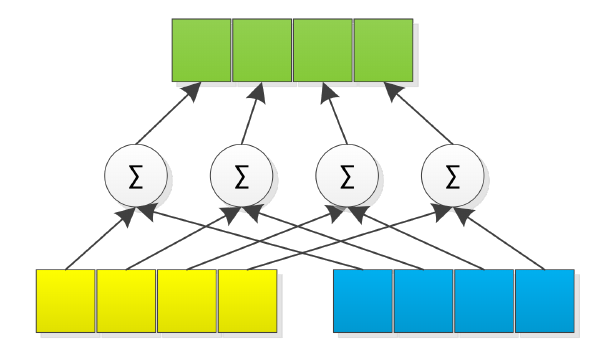}
			\caption{Score Concatenation with Score Merge}
			\label{fig:score_merge}
		\end{subfigure}
		\caption{A depiction of the four main data joining styles with two modalities.}
		\label{fig:data_fusion_style}
	\end{figure}

	During our review, we have identified four main styles of how data is joined during the fusion stage: Feature Concatenation, Deep Concatenation, Deep Merge, Score Concatenation. Figure \ref{fig:data_fusion_style} also provides visual representations of these commonly used styles. \newline
	
	\noindent \textbf{Feature Concatenation:} Features from multiple modalities are concatenated into a single feature vector as shown in Figure \ref{fig:feature_concat}.
	
	\noindent \textbf{Deep Concatenation:} The last layer of the modality specific sub-networks are simply pushed to a single layer without any other changes. Although the only difference between this and Feature Concatenation is the use of deep learning nodes, we believe this label helps to further identify the type of model being used as illustrated in Figure \ref{fig:deep_concat}.
		
	\noindent \textbf{Deep Merge:} This approach joins modality specific layers with rules more complicated than simple concatenation. Figure \ref{fig:deep_merge} shows the processes of going from eight nodes down to five with output nodes taking input from multiple modalities. Although the Deep Merge data joining style could be considered as concatenation followed by other neural network layers, some prior works have applied specific rules to this process and so it may be worthwhile to indicate the use of explicit business logic at this step. The Deep Merge style is often used with models employing some form of encoding.

 	\noindent \textbf{Score Concatenation:} The probability scores from multiple sub-classifiers are combined by either concatenation into a new feature vector or stacking the values in a matrix for ensemble scoring. These \textit{score features} can be either passed directly to traditional classifier or used to build a single score vector as shown in Figure \ref{fig:score_merge}. This example shows the Score Concatenation paired with a Score Merge classifier where an arithmetic operator is used for calculating the final score vector.	
	
	\subsection{Learning}
	\label{sec:learning}
	The learning stage is when the primary learning is performed, which can be done in one (early fusion) or more places (cross-modality and late fusion). These algorithms can include classifiers, autoencoders, CNNs, and other models. The learning stage is often simultaneously performed with the feature extraction or classification stages.

	\subsection{Classification}
	\label{sec:classifier}
	In unimodal learning or early fusion architectures, the classification is frequently done once. With late or cross-modality style models, classifiers are often used independently for each modality. Specific classifiers may be chosen for each modality but can be the same for all modalities. As an example, time series data may use an LSTM while image data may use a CNN. The output layers of these networks can then be merged using one of the data joining styles for the final classification which is often performed with a traditional machine learning classifier or a shallow DNN.
	
	\subsection{Multi-Stage Model Sharing}
\begin{figure}[ht]
		\centering
		\includegraphics[height=4cm]{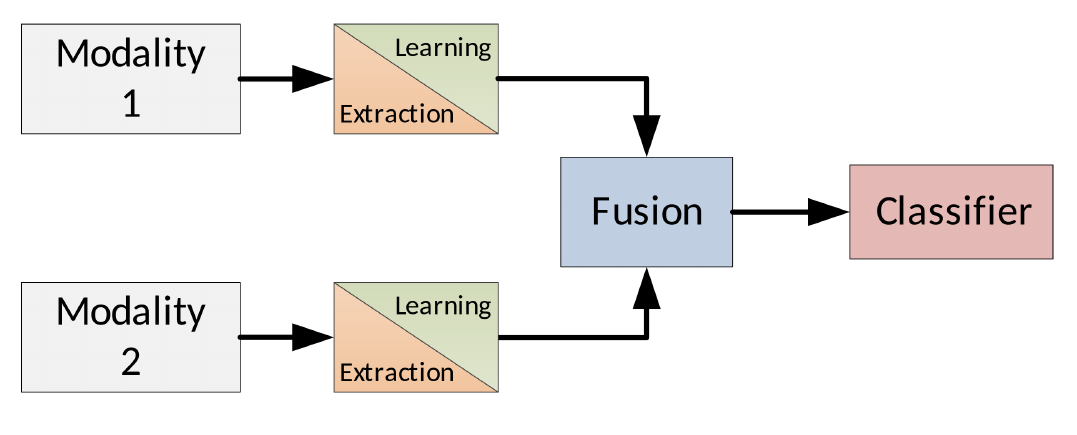}
		\caption{An example multimodal architecture using late fusion with the feature extraction and learning tasks performed with a shared model.}
		\label{fig:mixed1}
	\end{figure}

\begin{figure}[ht]
		\centering
		\includegraphics[height=3.75cm]{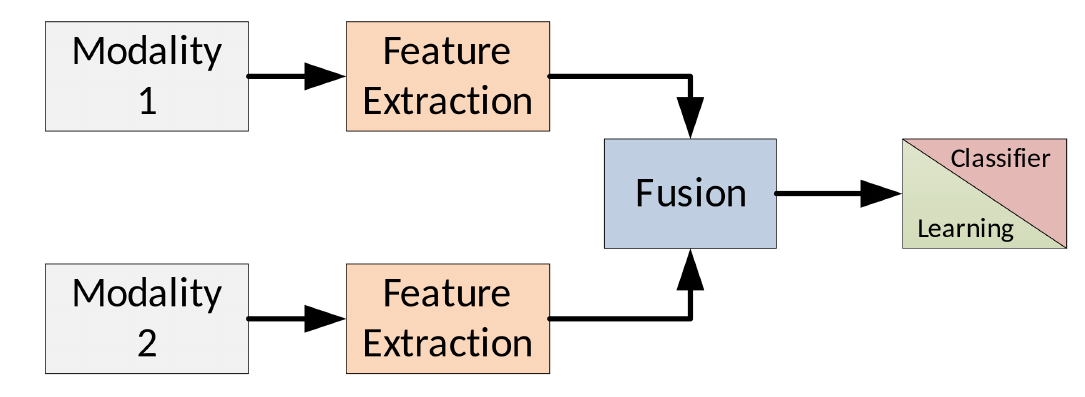}
		\caption{An example multimodal architecture using early fusion with the learning and classification tasks performed with a shared model.}
		\label{fig:mixed2}
	\end{figure}

	There are many real world scenarios where individual stages in our taxonomy use the same model. Figure \ref{fig:mixed1} shows an example of a late fusion architecture where each modality performs the feature extraction and learning with the same model. This is often done with CNNs as later shown in Table \ref{table:deeplearning}. Figure \ref{fig:mixed2} shows an early fusion architecture where the learning and classification is performed with a single model. This is done in the majority of traditional machine learning models as extracted features are concatenated and passed to a single classifier.
	
	\subsection{Applying the Taxonomy}
	While depictions of early and late fusion styles have been relatively consistent across multiple papers \cite{dahne2015multivariate, simonetta2019multimodal, ramachandram2017deep, baltruvsaitis2018multimodal, henderson2020improving, huang2020fusion, zhang2020deep}, there are still cases where other terms have been used. In \cite{guo2019deep}, one network architecture described as \textit{multi-view-one-network} is essentially early fusion and \textit{one-view-one-network} could be considered late fusion. In \cite{li2018review}, the authors provided a view-wise feature extraction architecture which could be mapped to our late fusion taxonomy with the use of different learner and classifier models. The authors of \cite{gao2020survey} described early fusion as \textit{shadow multiple modality} and late fusion as \textit{deep multiple modality}. For the cross-modality style architectures, \cite{ramachandram2017deep} and \cite{syed2021multi} used the term \textit{intermediate}, while \cite{gao2020survey} used \textit{deep shared modality}. In \cite{gao2020survey}, the authors also used \textit{deep cross modality} to describe multitask architectures but these terms were used in the context of generative models such as restricted Boltzmann machines (RBM). Even more specific models were described by \cite{oloyede2016unimodal} including \textit{Fusion at the Decision Level} (late fusion, score fusion), \textit{Fusion at the Matching Score Level} (late fusion, using different learners/classifiers), \textit{Biometric Traits at the Sensor Level} (early fusion), and \textit{Fusion at the Feature Level} (late fusion, feature concatenation). In a similar manner, \cite{yaman2019multimodal} used the terms \textit{data fusion} (early fusion), \textit{feature fusion} (late fusion, deep merge), and \textit{score fusion} (late fusion, score merge).
	
	The term \textit{intermediate} has been used for cross-modality fusion but can also refer to fusion occurring somewhere between feature extraction and classification \cite{choi2019embracenet}. Terms like \textit{middle} \cite{hong2020more}, \textit{joint} \cite{huang2020fusion}, and \textit{hybrid} \cite{zhang2020deep} have also been used to describe this kind of fusion. Since our proposed taxonomy is based on the five processing stages, the middle fusion concept can be captured by early or late fusion style architectures. This allows for the cross-modality fusion to describe only the inter-modality data sharing concept.
	
	These examples only cover a small portion of all existing works and so it is likely that many other depictions of multimodal classification architectures exist. Using our proposed taxonomy, we are able to label a wide range of models with consistent terms. In the following section, we review a number of multimodal classification models using this same taxonomy.

	\section{Review of Recent Classification Research}
	\label{sec:classification}
	
	In this section, we review recent works on multimodal classification. Tables \ref{table:machinelearning} and \ref{table:deeplearning} provide a selection of models published since 2017 and their architectural design using the previously defined taxonomy. In cases where a paper presented multiple models, we chose the one which gave the best overall results and for clarity some model types like CNNs were given their generic name instead of their specific implementation or the pre-trained model used. Determining the exact model configuration of these models may be open to different interpretations but we chose descriptions that best fit our proposed definitions. In these tables, the \textsuperscript{*} marker is used to identify cases where two of the stages are shared by the same model.

	\subsection{Traditional Machine Learning}
	\label{sec:ml_classifiers}

	\subsubsection{Early Fusion}	
	\label{sec:ml:early}

	Since many of the traditional machine learning classifiers are susceptible to the Curse of Dimensionality \cite{bellman1957dynamic}, explicit feature extraction is often performed for high dimensional data. This approach allows for the use of 2-D and 3-D imaging data which shows up often in the recently proposed multimodal models. For example, \cite{usman2017brain} used the Random Forest classifier with extracted imaging features from four MRI modalities to predict brain tumor status and \cite{zhou2019machine} also used MRI based features for genotyping brain tumors. \cite{kautzky2020machine} developed an attention-deficit and hyperactivity disorder (ADHD) diagnostic tool by using 49 region of interest (ROI) features with positron emission tomography (PET) images and 30 single nucleotide polymorphisms (SNP) based features. The authors of \cite{syed2021multi} built a predictive model for the automatic standardization of Digital Imaging and Communications in Medicine (DICOM) structure sets used in the Radiation Oncology field. The 3-D representation of each delineated structure was reduced to 50 features using singular value decomposition (SVD) and combined with word embedding features from the associated text annotations. Both sets of feature vectors were concatenated before using with the Random Forest classifier. Random Forest was also used with data from automated external defibrillators which provided electrocardiogram (ECG), thoracic impedance (TI), and the capnogram information \cite{elola2020multimodal}. This proposed model was time series based and manual feature engineering was performed before data fusion. \cite{panda2020multimodal} used electroencephalogram (EEG) and customer review text to predict emotional response. Data from both modalities were encoded into a shared feature space before being passed to the Random Forest classifier.

	Although Random Forest was the most commonly used traditional classifier, other algorithms were shown to be effective. For example, \cite{li2020multi} extracted 396 CT and MRI radiomic features from patients with locally advanced rectal cancer using the Linear Regression classifier for predicting therapeutic response after neoadjuvant chemotherapy. With the Gradient Boost Tree (GBT) classifier, \cite{liang2019classification} predicted individuals with schizophrenia using structural magnetic resonance imaging (sMRI) and diffusion tensor imaging based features. The authors of \cite{li2018hierarchical} developed a Support Vector Machine (SVM) based multimodal model for human activity and fall detection using an inertial measurement unit (IMU) and radar. However, this work used a hierarchical classification approach where sub-groups of features were grouped for predicting sub-activities. Instead of using a single classifier, \cite{huddar2018ensemble} built an ensemble of five individual classifiers. This work used transcription, audio, and video data for which a significant amount of preprocessing and feature selection was performed. Because the primary learning was done with the fused multimodal data, we have chosen to treat this as an early fusion architecture. 
	
	Using the Extreme Learning Machine (ELM) classifier, \cite{qureshi2017multimodal} built a model to assist in the diagnosis of schizophrenia. Features extracted from structural and functional MRI scans were grouped together to create new data modalities which were trained individually. Based on the their respective predictive power, each modality was given a weight used at the final multimodal classification step. \cite{gupta2020classification} created a predictive model for distinguishing healthy patients from those with Alzheimer's disease or mild cognitive impairment where transformed image and APOE genotype based features were shaped into kernel space before classification with the multiple kernel learning (MKL) algorithm \cite{aiolli2015easymkl}. \cite{sorinas2020brain} also used the \textit{k}-Nearest Neighbor (KNN) classifier with manually extracted features from EEG, ECG, and skin temperature data to predict emotional response to videos.

	
	\subsubsection{Late Fusion}	
	\label{sec:ml:late}
	In the work by \cite{guggenmos2020multimodal}, five neuroimaging modalities were used for classifying psychiatric disorders. Each modality was trained independently using both SVM and the weighted robust distance (WeiRD) classifiers. This model also used an optimization approach where the best classifier and hyperparameters were chosen for each modality. The final classification was performed using a weighed average of the results from each modality specific classifier. \cite{lee2019machine} also used SVMs for feature extraction and classification for predicting clinical pain states using brain imaging and heart rate data. To differentiate between healthy patients and those with either Alzheimer’s disease or mild cognitive impairment, \cite{lin2020predicting} developed a predictive model using MRI, fluorodeoxyglucose positron emission tomography (FDG-PET), cerebrospinal fluid (CSF), and apolipoprotein E (APOE) $\epsilon$4 gene data. Unlike the previous two methods, this model used the ELM classifier \cite{huang2011extreme} which is a variant of the traditional SVM algorithm.
	
	\subsubsection{Cross-modality Fusion}	
	\label{sec:ml:intermediate}	
	Using a multi-layer model with XGBoost, \cite{uddin2020multimodal} built a predictor for the presence of chronic back pain as part of the EmoPain 2020 Challenge \cite{aung2015automatic}. At each layer, some aspect of back pain was classified and the resulting class probabilities were then merged with the existing feature vectors. This fusion method shares some similarity to both early and late fusion architectures but has the unique property of progressively updating the feature vectors.
		
	\begin{table}[ht]
		\centering
		\renewcommand{\arraystretch}{1.25}
		\begin{adjustbox}{center,max width = 135mm}
			\footnotesize
			\begin{tabular}{lcccccl}\hline
				\multicolumn{7}{c}{\textbf{Multi-modal Classification with Traditional Machine Learning Techniques}}\\
				\bottomrule
				\textbf{Reference} & \textbf{Extraction} & \textbf{Fusion Type} & \textbf{Joiner} & \textbf{Learner} & \textbf{Classifier} & \textbf{Modalities} \\
				\bottomrule
				Usman and Rajpoot (2017) \cite{usman2017brain} & manual & early & feature concatenation & RF\textsuperscript{*} & RF\textsuperscript{*} & image \\
				Huddar et al. (2018) \cite{huddar2018ensemble} & manual/WE & early & feature concatenation & ensemble & score merge & audio/text/video \\
				Li et al. (2018) \cite{li2018hierarchical} & SVM\textsuperscript{*} & early & feature concatenation & SVM\textsuperscript{*} & SVM & signal\\
				Liang et al. (2019) \cite{liang2019classification} & manual & early & feature concatenation & GBT\textsuperscript{*} & GBT\textsuperscript{*} & image \\
				Elola et al. (2020) \cite{elola2020multimodal} & manual & early & feature concatenation & RF\textsuperscript{*} & RF\textsuperscript{*} & signal\\
				Kautzky et al. (2020) \cite{kautzky2020machine} & manual & early & feature concatenation & RF\textsuperscript{*} & RF\textsuperscript{*} & image/tabular \\
				
				Li et al. (2020) \cite{li2020multi} & manual & early & feature concatenation & LR\textsuperscript{*} & LR\textsuperscript{*} & image \\
				
		        Panda et al. (2020) \cite{panda2020multimodal} & manual & early & feature concatenation & RF\textsuperscript{*} & RF\textsuperscript{*} & tabular/text \\

				Syed et al. (2021) \cite{syed2021multi} & SVD/WE & early & feature concatenation & RF\textsuperscript{*} & RF\textsuperscript{*} & image/text \\
			

				Sorinas et al. (2020) \cite{sorinas2020brain} & manual & early & feature concatenation & KNN\textsuperscript{*} & KNN\textsuperscript{*} & signal \\
				
                Zhou et al. (2019) \cite{zhou2019machine} & manual & early & feature concatenation & RF\textsuperscript{*} & RF\textsuperscript{*} & image \\
                
                Gupta et al. (2020) \cite{gupta2020classification} & manual & early & feature concatenation & MKL & MKL & image/tabular \\
                
                Qureshi et al. (2017) \cite{qureshi2017multimodal} & manual & early & feature concatenation & ELM\textsuperscript{*} & ELM\textsuperscript{*} & image \\

                & & & & & & \\ 

                Lee et al. (2019) \cite{lee2019machine} & SVM\textsuperscript{*} & late & score concatenation & SVM\textsuperscript{*} & SVM & image/tabular \\
                
        		Guggenmos et al. (2020) \cite{guggenmos2020multimodal} & manual & late & score concatenation & SVM & score merge & image \\
                
                Lin et al. (2020) \cite{lin2020predicting} & manual & late & score concatenation & ELM & ELM & image/tabular \\
                & & & & & & \\ 
            	Uddin and Canavan (2020) \cite{uddin2020multimodal} & XGBoost & cross-modality & feature concatenation & XGBoost & XGBoost & tabular \\
				\bottomrule
			\end{tabular}
		\end{adjustbox}
		\caption{An overview of papers using traditional machine learning methods for multimodal classification.}
		\label{table:machinelearning}
	\end{table}

	\subsection{Deep Learning}
	\label{sec:dl_classifiers}
	
	\subsubsection{Early Fusion}	
	\label{sec:dl:early}
	
	To differentiate between patients with Mild Cognitive Impairment, Alzheimer’s disease, or without a neurological condition, \cite{ieracitano2020novel} used both Continuous Wavelet Transform (CWT) and bispectrum (BiS) data from EEG recordings. The experimental results showed that MLP outperformed AE, LR, and SVM classifiers with these concatenated features. Using both application-specific integrated circuit (ASIC) and field-programmable gate array (FPGA) based hardware platforms, \cite{jafari2018sensornet} used time series data from wearable sensors to detect human activity with a CNN. The sensor data was converted into a single channel image with one row per sensor and each column as a time series point. Using visible light and hyperspectral imaging, \cite{garillos2021multimodal} created a multimodal network for classifying papaya fruit maturity. The resulting RGB and hyperspectral image data was stacked as separate columns before being trained using a CNN. The authors of \cite{li2019multimodal} used multimodal satellite imagery to predict land cover types. Preprocessing was performed on each image modality and the resulting pixels were stacked into a single input vector for a DBN. Classification was then performed using SVM with the learned deep features.

	\subsubsection{Late Fusion}	
	\label{sec:l:late}
	

	Biometric identification systems can improve security with multiple verification methods. In the work by \cite{vijay2018multimodal}, ear and palm vein images were used and feature extraction was performed with the Multi-Support Vector Neural Network (Multi-SVNN) classifier for each modality. The final identification check was performed with the sum of the modality specific scores after optimizing the model weights. This work was later extended in \cite{vijay2021deep} using finger knuckle, ear, and iris image data. Although Mutli-SVNN was used again for learning each modality, a DBN was used for classification. Authors of \cite{alay2020deep} also built a biometric classifier with iris, face, and finger vein images. All three modalities were trained with CNNs and their output were passed through a softmax layer. These values were normalized before being combined using either an arithmetic mean or product rule and the highest value was chosen for the predicted class.	

	In \cite{aceto2019mimetic}, the authors developed a multimodal framework called MIMETIC for classifying network traffic. From each modality, data was extracted from network traces and independently trained with a CNN or gated recurrent unit (GRU) model. Final layers were concatenated and classified with DNN layers. This work was later extended to support multi-task classification with a new framework called DISTILLER \cite{aceto2021distiller}. This approach trained on shared DNN layers after the initial modal concatenation but the features were split again for training task specific layers.
	
	The field of medicine includes data from many potential sources including imaging, textual notes, and discrete information making it a natural domain for multimodal learning.
	In \cite{said2017multimodal}, an AE network was built for classifying EEG and electromyography (EMG) data. Each modality had its own AE which were merged and classification was performed by fine tuning a softmax function as the network bottleneck. Authors of \cite{tan2017multimodal} predicted cognitive events from EEG and optical flow temporal data. Both modalities are trained with CNNs and reshaped into a 2-D feature vector before being classified with a RNN. In \cite{venugopalan2021multimodal},  MRI, SNP, and clinical data were used for predicting cognitive disorders. The modalities were trained with CNN and AEs with their output layers concatenated and classified with a two layer DNN.
	
	Using cardiovascular and actigraphy sensing, \cite{zhai2020making} built an ensemble model for predicting sleep cycles. Three different time windows were chosen for both sensor types and the combined data was trained on CNN and LSTM models respectively, resulting in an ensemble of six total classifiers. All posterior probabilities were added to a classification matrix and the final classification was made with the highest average or argmax value. Using three derived image modalities from ECG data, \cite{ahmad2021ecg} built a model for heart beat classification. Each modality was trained on a CNN, the results were summed, and a SVM was chosen for the final classification. Clinical and cough audio data were used in \cite{agbley2020wavelet} to predict COVID-19 infections. The audio data was converted to a 2-D scalogram and trained with a CNN, while the clinical data was encoded using a DNN. The final layers were merged and classification was performed with a dense layer and softmax.

	In \cite{liu2021hybrid}, genomic data and pathology images were used to predict breast cancer subtypes. The genomic data was trained with a DNN and the image data with a CNN after principal component analysis (PCA) performed feature reduction. Authors of \cite{song2021multiview} used two image modalities from contrast-enhanced spectral mammography (CESM) scans for breast cancer detection. Each modality was trained with a CNN and the final layers were concatenated for classification with two DNN layers. For skin legion classification, \cite{yap2018multimodal} combined CNN generated features with tabular clinical data and performed classification with a three layer DNN. In \cite{ma2019brain}, MRI and pathology images were used to predict the cancer stage of brain tumors. Both modalities were classified with CNNs and their final layers were concatenated and classified using a LR classifier.
	 
	Using ground based cloud cover images and weather information, \cite{liu2018Amultimodal} created a joint fusion convolutional neural network (JFCNN). Image data was trained with a CNN and the weather data was trained with a decoder style DNN for feature learning. The final layers were concatenated and classified with a joint DNN layer. JFCNN was later used with GAN-generated artificial examples to increase the training dataset size \cite{liu2018multimodal}. In \cite{suzuki2018forest}, the authors used airborne imagery and geospatial features to classify forest cover. Each modality was trained with a CNN and the results were classified with CNN and DNN layers. Authors of \cite{xu2021mm} used image and tabular visit data for urban functional area classification. Learned features from the image and area visit information were trained with a GBT and their class probabilities were classified with a softmax layer.
	
	The authors of \cite{chancellor2017multimodal} used text and image data to detect pro-eating disorder policy violations on Tumblr. The text was trained using tag embedding with a fully connected layer and image data with a CNN. The resulting layers were concatenated and classified with a two layer DNN. In \cite{illendula2019multimodal}, a model was built to predict emotion from social media posts using image and text. The image data was classified with ResNet \cite{he2016deep}, the textual data with BiLSTM, and the final classification was performed with a softmax layer. In \cite{syed2021prediction}, audio, text, and image data were used to predict levels of public trust in politicians. Using CNN and word embedding models for each modality, the final classification was performed using either majority vote or the summed confidence scores.
		
	In \cite{gallo2017multimodal}, real world objects were predicted using image and text tags. A CNN was used to learn image features and the bag-of-words method was used for text. Experiments showed that SVM with images and Random Forest for text lead to best performance. Visual and near-infrared spectroscopy images were used in \cite{erickson2020multimodal} for helping robots interact with objects. The image data was trained with a DNN and the final layers were concatenated before classification with another two layer DNN.

	The authors of \cite{vielzeuf2017temporal} used video frames and audio data to classify the expressed emotion in video clips. Using CNN and RNN based networks, the final values for each modality were scored with a weighted mean. In \cite{oramas2018multimodal}, audio and album cover art were used to predict music genres. Each modality was trained with a CNN and two DNN layers, with the final results classified with a cosine loss function. The authors of \cite{tian2019multimodal} used audio, video frames, and text descriptions to identify different types of natural disasters. Feature extraction was performed with AENet \cite{takahashi2017aenet} for audio, Inception v3 \cite{szegedy2016rethinking} for video data, and text embedding with GloVe \cite{pennington2014glove}. The audio and video features were trained using an LSTM with the SVM-based Sequential Minimal Optimization (SMO) algorithm, and the text features were trained using a 1-D CNN to be classified with a softmax layer. Because some video concepts were not present in many of the video frames, the textual model was used for less common concepts instead of the joint audio-video.

	A DNN multimodal network was built in \cite{kang2017prediction} to predict crime occurrences with spatial, temporal, and environmental data. Each modality was trained with a DNN and the final layers were concatenated and classification was performed using two dense layers with softmax. Using text and the visual presentation of Wikipedia documents, \cite{shen2019joint} built a model to predict document quality. Text data was classified using BiSTLM and image data with Inception v3 for the image data. Final results were combined using a dense layer and softmax. Authors of \cite{zehtab2021multimodal} predicted smartphone prices with images of the devices and their discrete properties. Each modality was trained with CNNs and the output layers were flattened, concatenated, and classified with three more DNN layers. The EmbraceNet \cite{choi2019embracenet} framework was designed to accept data from any kind of input modality. This method embeds data from modality specific networks in a common length vector using \textit{docking} layers. A single \textit{embraced} vector is built from the \textit{docking} output using multinomial distribution so that each feature is only populated by a single modality.

	A different approach to multimodal learning is to create different feature sets from the same training data. For example, \cite{liang2021fusion} represented text as different modalities by splitting the data into phrases, words, n-grams, or other granularities using the Spatial View Attention Convolutional Neural Network (SVA-CNN) framework. This architecture was designed for preserving the relationships between these textual representations. Using context attention, parallel connection, and serial connection CNN sub-networks, the output convolutional layers were concatenated and classified with a DNN. Similarly, \cite{felipe2021automatic} also created several data modalities from Enteric Nervous System images of rats. Using a combination of handcrafted and model generated image features, several chronic degenerative diseases were classified. 

	\subsubsection{Cross-modality Fusion}
	\label{sec:dl:intermediate}

	Using both image and discrete weather station data, \cite{liu2019hierarchical} built a system to predict cloud types. Low level features were learned with a CNN for images and a fully connected network for the discrete data. These features were combined for further learning with another fully connected network and the resulting new features were combined with prior learned features before final classification. This network design included a multimodal skip connection, similar to what is found within neural network architectures like ResNet \cite{he2016deep}.

	The authors of \cite{yu2019entity} built a classification network for social media based sentiment analysis using image and text. First, an LSTM with average pooling was used to extract target entity information from the text. These results were combined with a CNN trained on the image data and with the textual context information trained on two other LSTMs. Several different combinations of fused features were concatenated for softmax classification. This approach allowed for different modalities to learn with information from the other modalities.
	
	 Different image modalities were used in \cite{hong2020more} to predict land cover and they tested multiple fusion network architectures. In addition to versions of early and late fusion, more advanced architectures were also investigated using both encoder-decoder and modality sharing schemes. The best results came from the latter methods which also used the deep merge data fusion style that compacted the multimodal data instead of concatenation.

	 A text and image based sentiment analysis network was built in \cite{yang2020image}. Text embedding and extracted image features were combined and trained on multi-modal CNN-LSTM based attention networks. Results from these networks were concatenated and classified with softmax to predict the emotion expressed in a social media post.

	For predicting brain diseases such as Alzheimer’s, \cite{gao2021task} used imaging data with a path-wise transfer network where partially extracted features were shared between modalities. At each layer of the network, weights from each modality were concatenated and convoluted before being concatenated again with the original modality outputs. This process allowed for the continuous sharing of information between each modality specific network. At the classification step, the final results from each modality were again concatenated and fine tuned with CNN, DNN, and softmax layers. In addition, this work also used a GAN to create artificial examples to address cases where one of the image modalities was missing.
	
	In one work \cite{huddar2020multi}, a multimodal network was created for sentiment analysis using audio, video, and text data. Feature selection was performed on each modality independently with a greedy search based genetic algorithm (GGA) followed by context extraction with a BiLSTM model. The unimodal results were then concatenated to form three new bimodal feature vectors representing the audio-video, text-video, and text-audio combinations. The same GGA/BiLSTM process was performed on those three modalities and the resulting vectors were again concatenated into a single feature vector. Finally, the combined vector was processed again with a GGA/BiLSTM followed by a softmax classifier.
	
	The Multi-Interactive Memory Network (MIMN) \cite{xu2019multi} was developed to predict sentiment labels from associated image and text information. Feature extraction was performed with word and phrase embedding for text data and a CNN for image data. Features from each modality were further processed their own LSTM before being used by two parallel memory networks, one for text and another for image data. Each network had multiple blocks that included a GRU and an attention mechanism. The first block accepts its matching image or text LSTM feature vector as well as the average pool from the aspect vector and returns the output of the GRU. In the following blocks, the input is the matching LSTM vector and the GRU output vector from the other parallel network, providing context from the other modality. The final classification is performed by concatenating the output of both networks followed by a softmax layer.

	
	\begin{table}[ht]
		\centering
		\renewcommand{\arraystretch}{1.25}
		\begin{adjustbox}{center,max width = 135mm}
			\footnotesize
			\begin{tabular}{lcccccl}\hline
				\multicolumn{7}{c}{\textbf{Multi-modal Classification with Deep Learning}}\\
				\bottomrule
				\textbf{Reference} & \textbf{Extraction} & \textbf{Fusion} & \textbf{Data Joiner} & \textbf{Learner} & \textbf{Classifier} & \textbf{Modalities} \\ 
				\bottomrule

				Jafari et al. (2018) \cite{jafari2018sensornet} & CNN\textsuperscript{*} & early & feature concatenation & CNN\textsuperscript{*} & DNN & signal \\
				Li et al. (2019) \cite{li2019multimodal} & manual & early & feature concatenation & DBN & SVM & image \\
				Ieracitano et al. (2020) \cite{ieracitano2020novel} & manual & early & feature concatenation & MLP\textsuperscript{*} & MLP\textsuperscript{*} & signal \\
				Garillos-Manliguez and Chiang (2021) \cite{garillos2021multimodal} & CNN\textsuperscript{*} & early & feature concatenation & CNN\textsuperscript{*} & DNN & image \\ 

				& & & & & & \\
				
				Chancellor et al. (2017) \cite{chancellor2017multimodal} & WE/CNN\textsuperscript{*} & late & deep concatenation & WE/CNN\textsuperscript{*} & DNN & image/text \\
				Gallo et al. (2017) \cite{gallo2017multimodal} & WE/CNN & late & score concatenation & SVM/RF & score merge & image/text\\
				Kang and Kang (2017) \cite{kang2017prediction} & DNN/CNN & late & deep concatenation & DNN & DNN & image/tabular \\
				
				Tan et al. (2017) \cite{tan2017multimodal} & CNN & late & deep concatenation & RNN & DNN & video \\
				Vielzeuf et al. (2017) \cite{vielzeuf2017temporal} & CNN/LSTM\textsuperscript{*} & late & score concatenation & CNN/LSTM\textsuperscript{*} & score merge & image/signal  \\
				Liu et al. (2018) \cite{liu2018Amultimodal} & DNN/CNN\textsuperscript{*} & late & deep concatenation & DNN/CNN\textsuperscript{*} & DNN & image/tabular \\
				Liu and Li (2018) \cite{liu2018multimodal} & CNN\textsuperscript{*} & late & deep concatenation & CNN\textsuperscript{*} & \textit{k}-NN & image/tabular \\
				Oramas et al. (2018) \cite{oramas2018multimodal} & CNN\textsuperscript{*} & late & deep concatenation & CNN\textsuperscript{*} & DNN & audio/image  \\
				Suzuki et al. (2018) \cite{suzuki2018forest}	& CNN & late & deep concatenation & CNN & DNN & image \\
				Vijay and Indumathi (2018) \cite{vijay2018multimodal} & manual & late & score concatenation & Multi-SVNN & score merge & image \\
				Yap et al. (2018) \cite{yap2018multimodal} & CNN & late & deep concatenation & DNN\textsuperscript{*} & DNN\textsuperscript{*} & image/tabular \\
				Aceto et al. (2019) \cite{aceto2019mimetic} & CNN/GRU\textsuperscript{*} & late & deep concatenation & CNN/GRU\textsuperscript{*} & DNN & tabular \\
				
				Choi et al. (2019) \cite{choi2019embracenet} & CNN & late & score merge & DNN\textsuperscript{*} & DNN\textsuperscript{*} & tabular \\
				
				Illendula and Sheth (2019) \cite{illendula2019multimodal} & CNN/BiLSTM\textsuperscript{*} & late & deep concatenation & CNN/BiLSTM\textsuperscript{*} & softmax & image/text  \\
				Jaiswal et al. (2019) \cite{jaiswal2019controlling} & CNN/GRU\textsuperscript{*} & late & deep concatenation & CNN/GRU\textsuperscript{*} & DNN & audio/text \\
				Ma and Jia (2019) \cite{ma2019brain}	& CNN\textsuperscript{*} & late & deep concatenation & CNN\textsuperscript{*} & LR & image \\
				Shen et al. (2019) \cite{shen2019joint} & LSTM/CNN\textsuperscript{*} & late & deep concatenation & LSTM/CNN\textsuperscript{*} & DNN & image/text \\
				Tian et al. (2019) \cite{tian2019multimodal} & AE/CNN/WE & late & deep merge & LSTM/WE & DNN/CNN & audio/image/text \\
				Xu et al. (2019) \cite{xu2019multimodal} & CNN/LSTM/DT\textsuperscript{*} & late & score concatenation & CNN/LSTM/DT\textsuperscript{*} & score merge & tabular/text  \\
				Agbley et al. (2020) \cite{agbley2020wavelet} & DNN/CNN\textsuperscript{*} & late & dense concatenation & DNN/CNN\textsuperscript{*} & DNN & tabular/signal \\
				Alay and Al-Baity (2020) \cite{alay2020deep} & CNN\textsuperscript{*} & late & score concatenation & CNN\textsuperscript{*} & score merge & image  \\
				Erickson et al. (2020) \cite{erickson2020multimodal} & CNN & late & deep concatenation & DNN & DNN & image \\
				Gadiraju et al. (2020) \cite{gadiraju2020multimodal} & CNN/LSTM\textsuperscript{*} & late & score concatenation & CNN/LSTM\textsuperscript{*} & SVM & image/signal \\
				Zhai et al. (2020) \cite{zhai2020making} & CNN/LSTM\textsuperscript{*} & late & score concatenation & CNN/LSTM\textsuperscript{*} & score merge & signal  \\
				Ahmad et al. (2021) \cite{ahmad2021ecg} & CNN & late & deep merge & CNN & SVM & signal \\
				Aceto et al. (2021) \cite{aceto2021distiller}& CNN/GRU & late & deep concatenation & CNN/GRU & DNN & tabular  \\
				
				Felipe et al. (2021) \cite{felipe2021automatic} & manual/CNN & late & score concatenation & CNN/SVM & score merge & image \\		
				Liang et al. (2021) \cite{liang2021fusion} & CNN & late & deep concatenation & CNN & DNN & text \\
				Liu et al. (2021) \cite{liu2021hybrid} & CNN/DNN\textsuperscript{*} & late & score concatenation & CNN/DNN\textsuperscript{*} & score merge & image/tabular \\
				Song et al. (2021) \cite{song2021multiview} & CNN\textsuperscript{*} & late & deep concatenation & CNN\textsuperscript{*} & DNN & image \\
				Syed et al. (2021) \cite{syed2021prediction} & CNN/WE & late & score concatenation & LR & score merge & audio/image/text \\
				Venugopalan et al. (2021) \cite{venugopalan2021multimodal} & CNN\textsuperscript{*}/manual & late & deep concatenation & DNN/CNN\textsuperscript{*} & DNN & tabular/image \\ 
				Vijay and Indumathi (2021) \cite{vijay2021deep} & manual & late & score concatenation & Multi-SVNN & DBN & image \\
				Xu et al. (2021) \cite{xu2021mm} & CNN & late & score concatenation & CNN/GBT & GBT & image/tabular \\
				Zehtab-Salmasi et al. (2021) \cite{zehtab2021multimodal} & CNN\textsuperscript{*} & late & deep concatenation & CNN\textsuperscript{*} & DNN & image/tabular \\
				
				& & & & & & \\		
				Said et al. (2017) \cite{said2017multimodal} & AE\textsuperscript{*} & cross-modality & deep merge & AE\textsuperscript{*} & DNN & signal\\
				Liu et al. (2019) \cite{liu2019hierarchical} & DNN/CNN & cross-modality & deep concatenation & DNN & DNN & image/tabular \\
				
				Xu et al. (2019) \cite{xu2019multi} & BiLSTM/CNN & cross-modality & deep concatenation & GRU & DNN & image/text \\
				
				Yu et al. (2019) \cite{yu2019entity} & CNN/LSTM\textsuperscript{*} & cross-modality & deep concatenation & CNN/LSTM\textsuperscript{*} & softmax & image/text \\
				Hong et al. (2020) \cite{hong2020more} & CNN & cross-modality & deep merge & CNN & softmax & image \\
				Huddar et al. (2020) \cite{huddar2020multi} & BiLSTM/GGA & cross-modality & ”tensor fusion” & BiLSTM & DNN & audio/text/video \\
				Yang et al. (2020) \cite{yang2020image} & CNN/WE & cross-modality & deep concatenation & AN & DNN/CNN & image/text \\
				Gao et al. (2021) \cite{gao2021task} & CNN\textsuperscript{*} & cross-modality & deep concatenation & CNN\textsuperscript{*} & DNN & image \\

				\bottomrule
			\end{tabular}
		\end{adjustbox}
		\caption{An overview of articles using traditional deep learning methods for multimodal classification.}
		\label{table:deeplearning}
	\end{table}

	\subsection{Observations}
	\label{sec:trends}
	
	From our literature search, it was clear that the current focus of multimodal classification is with deep learning although new research based on traditional machine learning is still being produced. Compared to deep learning, the traditional machine learning models primarily use the manual style feature extraction. The early fusion type was used in 13 of the 17 traditional machine learning models, likely because these classifiers expect a single feature vector or matrix as input and does not provide a way to further augment the data during training. For the same reason, those models often used simple multimodal feature concatenation. Most works also used the same algorithm for learning and classification unless it was a multi-task problem or used an ensemble. Tree base algorithms, such as Random Forest, GBT, and XGBoost, were the most popular classifiers. All of the major data types were used in the machine learning models but images were used most often.
	
	In the deep learning papers, four used early fusion, eight used cross modality fusion, and 35 used late fusion, a reversal from the traditional machine learning works. Feature extraction was usually data type dependent but CNN was by far the most popular method followed by RNNs, such as LSTM and GRU. The architectural design of the deep learning early fusion models were similar to those using machine learning, with the biggest difference being the individual algorithms used. Feature concatenation was used in all four cases even if CNN was used for feature extraction. 

	Models using late fusion perform the bulk of modality training independently which allows for the use of pre-trained models in deep learning solutions. This also supports the use of fundamentally different models for each specific modality, such as CNN for images and LSTM for text. The majority of late fusion methods concatenated or merged the final layer of the modality specific networks as the data joining method. This is a straightforward way to transition the partial results into the classifier layers. However, class probabilities from the modality specific classifiers were also combined with an arithmetic function or were concatenated to a new feature vector for the final classification. In most late fusion models, a shallow DNN followed by a softmax layer was used as the classifier. 
	
	While cross-modality fusion is still not the most common method, its popularity may be increasing as it was more prominent in the deep learning models. These architectures tend to be more complicated than early or late fusion, but it has been shown that the performance may be superior. At this time, multimodal learning lacks large pre-trained networks that are available for unimodal learning such as ResNet, Inception v3, or VGG \cite{simonyan2014very}. If the best performing cross-modality architectures could be pre-trained on very large datasets, the resulting models would provide a significant benefit for future work.
	
	Several sub-network patterns have also emerged from these prior works. The most common architecture was a shared CNN for feature extraction and modality learning, followed by a DNN classifier. In cases where different network types were used for each modality, CNN and RNN networks were often paired.

	Architectures using the learning stage models as full classifiers were more likely to use score merge as the final classification method instead of a DNN. This approach may have been often used because it is convenient to simply apply a final softmax layer to the pre-trained model outputs. However, future work is needed to determine if that method is superior to merging the modality specific networks with the deep merge style data fusion.



	\section{Discussion of Open Problems}
	\label{sec:limits}
	While there has been much progress in the recent years with multimodal data classification, there are still several important areas that have not been adequately addressed. In this section, we discuss open problems related to the ever growing size of datasets, difficult classification tasks, and the lack of general tools for multimodel classification.
	

	\subsection{Big Data}
	The emergence of Big Data has led to new opportunities and challenges. While there have been many effective solutions for classification on large data sets, little has been done specifically for multimodal data as unimodal has been the focus up to this point. One of the major limitation with multimodal learning research is the lack of large, publicly available datasets. The vast majority of experiments carried out in the reviewed papers used small datasets with only tens to thousands of examples. The private datasets, often containing medical imaging, also were small as collecting healthcare data can be time consuming, expensive, and often comes with ethical and legal restrictions on its use.

	Deep learning networks are currently the most popular way of performing multimodal classification and this technology works best with large training datasets. Although \cite{illendula2019multimodal} used a dataset with approximately 500,000 examples, the majority of other works used significantly fewer. The challenge of limited data is further compounded when presented with multiclass datasets, as the total number of examples per class will be even further reduced. The creation of large benchmark datasets which covers different modality combinations will be a significant benefit for future multimodal research.
		
	While the lack of data may negatively affect these models, its impact on multimodal classification has not been fully investigated. Transfer learning with pre-trained models may help with limited training data and this approach has been used in a number of the reviewed papers. Data augmentation is a common method for inserting new training examples such as image shifts or rotations. GANs have also been used to add synthetic examples to increase the size of multimodal datasets \cite{liu2018multimodal, li2020deep}. 
	
	Large datasets are often not fully curated and can have missing or erroneous values. Since these issues can exist independently for each modality, it is likely that a higher percentage of training examples will require data cleaning. In additional to traditional data imputation, it was also shown that GANs can be used to add data missing from one of the modalities \cite{gao2021task}. However, even if GAN-generated or cleaned examples appear to be realistic, it is important to ensure that inter-modality relationships are also representative of the original dataset. Future work is needed to provide methods that discover these relationships and how the training data can be safely modified.

	\subsection{Parallel and Distributed Computing}
	The relatively small datasets used in current multimodal research has not required distributed computing solutions. However, classification models on larger datasets will need more resources than are available with a single CPU or GPU. Distributed systems like Apache Hadoop, Apache Spark, and GPU clusters are potential solutions for training large datasets but their current tools are not designed for multimodal learning. In addition to the technical expertise required for implementing such parallel systems, the distribution of data between computational nodes may also impact model performance. If the distribution of class instances or sub-concepts are not carefully considered, partial results generated at each node may not properly reflect the global properties of the training data \cite{sleeman2021multi}. While this has been shown to be a potential issue with unimodal learning, it has not yet been shown how this would affect multimodal models.
	
	\subsection{Imbalanced Data}
	Many machine learning algorithms were designed with the assumption that class distributions are balanced in the training dataset. However, this is often not the case and class imbalance may introduce a bias that could negatively affect classifier performance \cite{he2009learning}. In many real world datasets, there may be a majority class that is significantly larger than the other minority class. The same problem is magnified with multiclass datasets as there may be many majority-minority class relationships.	
	
	Class imbalanced data can be addressed by modifying the data itself, using algorithms inherently designed for addressing this issue, or a combination of both \cite{krawczyk2016learning}. Data level methods have been most popular with unimodal datasets and was the only method found within the multimodal models included in this literature review. Random oversampling is one of the most common data level methods, where random examples are simply replicated until the desired level of class balance is achieved. Random undersampling has been used but may run the risk of removing useful examples or producing a final training dataset that is too small overall. Another popular method is the synthetic minority over-sampling technique (SMOTE) \cite{chawla2002smote} which generates new examples from a combination of nearby examples of the same class. This approach has inspired the development of many other oversampling algorithms and has been shown to outperform random sampling methods in many cases.
	
	While there has been limited work done on multimodal imbalanced learning, it was addressed in a few of the reviewed papers. For example, \cite{kang2017prediction} and \cite{usman2017brain} used random undersampling while \cite{elola2020multimodal} used random oversampling. The more advanced SMOTE method was also used by \cite{li2020multi} and \cite{uddin2020multimodal}. Since SMOTE is based on the \textit{k}-NN algorithm, it is important that the individual features are normalized or weighed by importance. Without performing this step, the generation of synthetic examples may be over influenced by noisy or less impactful features and thus resulting in lower quality examples. 
	
	GANs have become a popular method for data augmentation \cite{shorten2019survey} and have already been used with multimodal datasets. In one work, \cite{pahde2021multimodal} used a GAN for adding missing text modality data which was paired with image data. Moreover, \cite{li2020deep} generated completely new synthetic examples instead of just imputing missing modality data. Future work is needed to better understand how methods like SMOTE and GANs could be best utilized with multimodal datasets.
	
	\subsection{Instance Level Difficulty}
	Poor classifier performance cannot always be blamed on class imbalance, especially if classes are well separated. Some examples are harder to learn from because they exist in a region contaminated with examples from other classes or are close to decision boundaries \cite{saez2016analyzing}. The difficulty of learning from such examples increases with more classes or a higher level of class imbalance, as both can lead to more regional contamination. These challenges have been well researched in traditional unimodal learning and a number of solutions have been proposed to address this issue, especially when presented with class imbalance.
	
	Popular oversampling methods such as random oversampling and SMOTE have been shown to do well at addressing class imbalance but do not consider which examples are more important for classification. Algorithms like Save Level SMOTE \cite{safe_level_smote} and Borderline SMOTE \cite{borderlineSMOTE} favor specific types of examples during oversampling to add more emphasis on certain parts of the feature space, such as safe class regions and decision boundaries. However, no such method have been designed that includes these learning concepts with multimodal datasets.
	
	These unimodal solutions may not directly translate to multimodal problems as individual modalities could exhibit different levels of learning difficulty. Embedding each modality to a shared latent space may help solve this problem but more experimentation is required to identify the best solutions.
	
	
	\subsection{Evaluation Metrics}
	Classification models must be evaluated to determine their performance and the specific metrics used are dependent on the desired results. True positive rate, precision, recall, and F$_{1}$ scores are often used, especially for binary class datasets. The macro and micro average of F$_{1}$ is also used for multiclass datasets but these metrics may provide over optimistic results when presented with imbalanced data as it may hide poor performance on minority classes. As reported by \cite{branco2017relevance}, there are a number of other metrics that are better suited for dealing with imbalanced data, such as Average Accuracy (AvAcc), the Macro Average Geometric (MAvG), and Class Balance Accuracy (CBA).

	These existing metrics were designed for unimodal problems and we are unaware of any metrics designed specifically for evaluating multimodal classifiers. While these metrics are addressing performance related to the predicted classes, there may be value in knowing how well each model does with each individual modality. Going further, the performance of each class may be affected differently by each modality. New metrics designed for multimodal classification could help identify areas in which a modality specific model is under performing as well as provide a better understanding of the relationships between classes and each modality.
	
		
	\subsection{Universal Models}
	As shown in previous research, traditional machine learning algorithms and deep learning architectures can be successfully used for multimodal classification. However, the general usability of these models is limited as most are tailored for their domain specific input modalities and may not work directly on different data combinations. Although the EmbraceNet \cite{choi2019embracenet} partially addresses this issue, it was only designed for late fusion style networks and may lose some shared context between modalities depending on how the fused feature vector is constructed. Ideally, future multimodal frameworks will be configurable to support arbitrary input types using well defined architectural rules. This will allow for a straightforward manual or automated construction of a complicated model or network. With the success of transfer learning and an ever increasing number of pre-trained models, providing plug-and-play support for these networks will simplify the process of constructing powerful multimodal models. 
	

	
	\section{Conclusions}
	\label{sec:con}
	
	Although unimodal learning has dominated the machine learning field, there is a growing interest in multimodal problems. New methods for combining data from multiple sources, the large collection of social media and customer reviews, and the aggregation of healthcare related information are all providing more valuable use cases for multimodal learning. The general consensus from the reviewed papers is that multimodal based architectures have the potential of outperforming the traditional unimodal models. However, many of the most difficult problems facing classification, such as big data, class imbalance, and instance level difficulty, have not yet been fully addressed in this context. We have also seen cases where unimodal datasets can be treated as multimodal problems and thus utilize these novel learning methods.
	
	There are a number of other important challenges not addressed in this paper that will require further research. Regression models have successfully been used for mutlimodal datasets \cite{wang2019knowledge, wang2018retweet, lee2019machine} but have gotten much less focus than classification. A future survey on this topic could help identify if the general challenges and our proposed taxonomy for classification translates well to regression models. Generative models, such as GANs and AEs, have been well represented in multimodal survey papers but were used in only a few of the reviewed classification models. In a similar manner, Canonical Correlation Analysis (CCA) has been commonly discussed in the context of multimodal learning but is less often used with classification models. An in-depth discussion of how or when to use such methods for multimodal classification would be a benefit to the research community.
	
	In summary, we have proposed a new multimodal classification taxonomy for describing both the overall model architectures and the style in which data fusion is performed. Unlike previous taxonomies, we focus solely on classification problems and guided our definitions based on common patterns from prior works. We believe this kind of taxonomy will be helpful when describing multimodal models which tend to be more complicated than their unimodal counterparts.


\bibliographystyle{ACM-Reference-Format}
\bibliography{references}


\begin{thebibliography}{121}


\ifx \showCODEN    \undefined \def \showCODEN     #1{\unskip}     \fi
\ifx \showDOI      \undefined \def \showDOI       #1{#1}\fi
\ifx \showISBNx    \undefined \def \showISBNx     #1{\unskip}     \fi
\ifx \showISBNxiii \undefined \def \showISBNxiii  #1{\unskip}     \fi
\ifx \showISSN     \undefined \def \showISSN      #1{\unskip}     \fi
\ifx \showLCCN     \undefined \def \showLCCN      #1{\unskip}     \fi
\ifx \shownote     \undefined \def \shownote      #1{#1}          \fi
\ifx \showarticletitle \undefined \def \showarticletitle #1{#1}   \fi
\ifx \showURL      \undefined \def \showURL       {\relax}        \fi
\providecommand\bibfield[2]{#2}
\providecommand\bibinfo[2]{#2}
\providecommand\natexlab[1]{#1}
\providecommand\showeprint[2][]{arXiv:#2}

\bibitem[\protect\citeauthoryear{Aceto, Ciuonzo, Montieri, and
  Pescap{\`e}}{Aceto et~al\mbox{.}}{2019}]%
        {aceto2019mimetic}
\bibfield{author}{\bibinfo{person}{Giuseppe Aceto}, \bibinfo{person}{Domenico
  Ciuonzo}, \bibinfo{person}{Antonio Montieri}, {and} \bibinfo{person}{Antonio
  Pescap{\`e}}.} \bibinfo{year}{2019}\natexlab{}.
\newblock \showarticletitle{MIMETIC: Mobile encrypted traffic classification
  using multimodal deep learning}.
\newblock \bibinfo{journal}{\emph{Computer Networks}}  \bibinfo{volume}{165}
  (\bibinfo{year}{2019}), \bibinfo{pages}{106944}.
\newblock


\bibitem[\protect\citeauthoryear{Aceto, Ciuonzo, Montieri, and
  Pescap{\'e}}{Aceto et~al\mbox{.}}{2021}]%
        {aceto2021distiller}
\bibfield{author}{\bibinfo{person}{Giuseppe Aceto}, \bibinfo{person}{Domenico
  Ciuonzo}, \bibinfo{person}{Antonio Montieri}, {and} \bibinfo{person}{Antonio
  Pescap{\'e}}.} \bibinfo{year}{2021}\natexlab{}.
\newblock \showarticletitle{DISTILLER: Encrypted traffic classification via
  multimodal multitask deep learning}.
\newblock \bibinfo{journal}{\emph{Journal of Network and Computer
  Applications}}  \bibinfo{volume}{183} (\bibinfo{year}{2021}),
  \bibinfo{pages}{102985}.
\newblock


\bibitem[\protect\citeauthoryear{Agbley, Li, Haq, Cobbinah, Kulevome, Agbefu,
  and Eleeza}{Agbley et~al\mbox{.}}{2020}]%
        {agbley2020wavelet}
\bibfield{author}{\bibinfo{person}{Bless Lord~Y Agbley},
  \bibinfo{person}{Jianping Li}, \bibinfo{person}{Aminul Haq},
  \bibinfo{person}{Bernard Cobbinah}, \bibinfo{person}{Delanyo Kulevome},
  \bibinfo{person}{Priscilla~A Agbefu}, {and} \bibinfo{person}{Bright Eleeza}.}
  \bibinfo{year}{2020}\natexlab{}.
\newblock \showarticletitle{Wavelet-based cough signal decomposition for
  multimodal classification}. In \bibinfo{booktitle}{\emph{2020 17th
  International Computer Conference on Wavelet Active Media Technology and
  Information Processing (ICCWAMTIP)}}. IEEE, \bibinfo{pages}{5--9}.
\newblock


\bibitem[\protect\citeauthoryear{Ahmad, Tabassum, Guan, and Khan}{Ahmad
  et~al\mbox{.}}{2021}]%
        {ahmad2021ecg}
\bibfield{author}{\bibinfo{person}{Zeeshan Ahmad}, \bibinfo{person}{Anika
  Tabassum}, \bibinfo{person}{Ling Guan}, {and} \bibinfo{person}{Naimul~Mefraz
  Khan}.} \bibinfo{year}{2021}\natexlab{}.
\newblock \showarticletitle{ECG Heartbeat Classification Using Multimodal
  Fusion}.
\newblock \bibinfo{journal}{\emph{IEEE Access}} (\bibinfo{year}{2021}).
\newblock


\bibitem[\protect\citeauthoryear{Aiolli and Donini}{Aiolli and Donini}{2015}]%
        {aiolli2015easymkl}
\bibfield{author}{\bibinfo{person}{Fabio Aiolli} {and} \bibinfo{person}{Michele
  Donini}.} \bibinfo{year}{2015}\natexlab{}.
\newblock \showarticletitle{EasyMKL: a scalable multiple kernel learning
  algorithm}.
\newblock \bibinfo{journal}{\emph{Neurocomputing}}  \bibinfo{volume}{169}
  (\bibinfo{year}{2015}), \bibinfo{pages}{215--224}.
\newblock


\bibitem[\protect\citeauthoryear{Alay and Al-Baity}{Alay and Al-Baity}{2020}]%
        {alay2020deep}
\bibfield{author}{\bibinfo{person}{Nada Alay} {and} \bibinfo{person}{Heyam~H
  Al-Baity}.} \bibinfo{year}{2020}\natexlab{}.
\newblock \showarticletitle{Deep Learning Approach for Multimodal Biometric
  Recognition System Based on Fusion of Iris, Face, and Finger Vein Traits}.
\newblock \bibinfo{journal}{\emph{Sensors}} \bibinfo{volume}{20},
  \bibinfo{number}{19} (\bibinfo{year}{2020}), \bibinfo{pages}{5523}.
\newblock


\bibitem[\protect\citeauthoryear{Audebert, Le~Saux, and Lef{\`e}vre}{Audebert
  et~al\mbox{.}}{2019}]%
        {audebert2019deep}
\bibfield{author}{\bibinfo{person}{Nicolas Audebert}, \bibinfo{person}{Bertrand
  Le~Saux}, {and} \bibinfo{person}{S{\'e}bastien Lef{\`e}vre}.}
  \bibinfo{year}{2019}\natexlab{}.
\newblock \showarticletitle{Deep learning for classification of hyperspectral
  data: A comparative review}.
\newblock \bibinfo{journal}{\emph{IEEE geoscience and remote sensing magazine}}
  \bibinfo{volume}{7}, \bibinfo{number}{2} (\bibinfo{year}{2019}),
  \bibinfo{pages}{159--173}.
\newblock


\bibitem[\protect\citeauthoryear{Aung, Kaltwang, Romera-Paredes, Martinez,
  Singh, Cella, Valstar, Meng, Kemp, Shafizadeh, et~al\mbox{.}}{Aung
  et~al\mbox{.}}{2015}]%
        {aung2015automatic}
\bibfield{author}{\bibinfo{person}{Min~SH Aung}, \bibinfo{person}{Sebastian
  Kaltwang}, \bibinfo{person}{Bernardino Romera-Paredes},
  \bibinfo{person}{Brais Martinez}, \bibinfo{person}{Aneesha Singh},
  \bibinfo{person}{Matteo Cella}, \bibinfo{person}{Michel Valstar},
  \bibinfo{person}{Hongying Meng}, \bibinfo{person}{Andrew Kemp},
  \bibinfo{person}{Moshen Shafizadeh}, {et~al\mbox{.}}}
  \bibinfo{year}{2015}\natexlab{}.
\newblock \showarticletitle{The automatic detection of chronic pain-related
  expression: requirements, challenges and the multimodal EmoPain dataset}.
\newblock \bibinfo{journal}{\emph{IEEE transactions on affective computing}}
  \bibinfo{volume}{7}, \bibinfo{number}{4} (\bibinfo{year}{2015}),
  \bibinfo{pages}{435--451}.
\newblock


\bibitem[\protect\citeauthoryear{Baltru{\v{s}}aitis, Ahuja, and
  Morency}{Baltru{\v{s}}aitis et~al\mbox{.}}{2018}]%
        {baltruvsaitis2018multimodal}
\bibfield{author}{\bibinfo{person}{Tadas Baltru{\v{s}}aitis},
  \bibinfo{person}{Chaitanya Ahuja}, {and} \bibinfo{person}{Louis-Philippe
  Morency}.} \bibinfo{year}{2018}\natexlab{}.
\newblock \showarticletitle{Multimodal machine learning: A survey and
  taxonomy}.
\newblock \bibinfo{journal}{\emph{IEEE transactions on pattern analysis and
  machine intelligence}} \bibinfo{volume}{41}, \bibinfo{number}{2}
  (\bibinfo{year}{2018}), \bibinfo{pages}{423--443}.
\newblock


\bibitem[\protect\citeauthoryear{Bellman, Corporation, and Collection}{Bellman
  et~al\mbox{.}}{1957}]%
        {bellman1957dynamic}
\bibfield{author}{\bibinfo{person}{R. Bellman}, \bibinfo{person}{Rand
  Corporation}, {and} \bibinfo{person}{Karreman Mathematics~Research
  Collection}.} \bibinfo{year}{1957}\natexlab{}.
\newblock \bibinfo{booktitle}{\emph{Dynamic Programming}}.
\newblock \bibinfo{publisher}{Princeton University Press}.
\newblock
\showISBNx{9780691079516}
\showLCCN{57005444}


\bibitem[\protect\citeauthoryear{Branco, Torgo, and Ribeiro}{Branco
  et~al\mbox{.}}{2017}]%
        {branco2017relevance}
\bibfield{author}{\bibinfo{person}{Paula Branco}, \bibinfo{person}{Lu{\'\i}s
  Torgo}, {and} \bibinfo{person}{Rita~P Ribeiro}.}
  \bibinfo{year}{2017}\natexlab{}.
\newblock \showarticletitle{Relevance-based evaluation metrics for multi-class
  imbalanced domains}. In \bibinfo{booktitle}{\emph{Pacific-Asia Conference on
  Knowledge Discovery and Data Mining}}. Springer, \bibinfo{pages}{698--710}.
\newblock


\bibitem[\protect\citeauthoryear{Bunkhumpornpat, Sinapiromsaran, and
  Lursinsap}{Bunkhumpornpat et~al\mbox{.}}{2009}]%
        {safe_level_smote}
\bibfield{author}{\bibinfo{person}{Chumphol Bunkhumpornpat},
  \bibinfo{person}{Krung Sinapiromsaran}, {and} \bibinfo{person}{Chidchanok
  Lursinsap}.} \bibinfo{year}{2009}\natexlab{}.
\newblock \showarticletitle{Safe-Level-SMOTE: Safe-Level-Synthetic Minority
  Over-Sampling TEchnique for Handling the Class Imbalanced Problem}. In
  \bibinfo{booktitle}{\emph{Proceedings of the 13th Pacific-Asia Conference on
  Advances in Knowledge Discovery and Data Mining}} (Bangkok, Thailand)
  \emph{(\bibinfo{series}{PAKDD '09})}. \bibinfo{publisher}{Springer-Verlag},
  \bibinfo{address}{Berlin, Heidelberg}, \bibinfo{pages}{475--482}.
\newblock
\showISBNx{978-3-642-01306-5}
\urldef\tempurl%
\url{https://doi.org/10.1007/978-3-642-01307-2_43}
\showDOI{\tempurl}


\bibitem[\protect\citeauthoryear{Chancellor, Kalantidis, Pater, De~Choudhury,
  and Shamma}{Chancellor et~al\mbox{.}}{2017}]%
        {chancellor2017multimodal}
\bibfield{author}{\bibinfo{person}{Stevie Chancellor}, \bibinfo{person}{Yannis
  Kalantidis}, \bibinfo{person}{Jessica~A Pater}, \bibinfo{person}{Munmun
  De~Choudhury}, {and} \bibinfo{person}{David~A Shamma}.}
  \bibinfo{year}{2017}\natexlab{}.
\newblock \showarticletitle{Multimodal classification of moderated online
  pro-eating disorder content}. In \bibinfo{booktitle}{\emph{Proceedings of the
  2017 CHI Conference on Human Factors in Computing Systems}}.
  \bibinfo{pages}{3213--3226}.
\newblock


\bibitem[\protect\citeauthoryear{Chandrasekaran, Nguyen, and
  Hemanth~D}{Chandrasekaran et~al\mbox{.}}{2021}]%
        {chandrasekaran2021multimodal}
\bibfield{author}{\bibinfo{person}{Ganesh Chandrasekaran},
  \bibinfo{person}{Tu~N Nguyen}, {and} \bibinfo{person}{Jude Hemanth~D}.}
  \bibinfo{year}{2021}\natexlab{}.
\newblock \showarticletitle{Multimodal sentimental analysis for social media
  applications: A comprehensive review}.
\newblock \bibinfo{journal}{\emph{Wiley Interdisciplinary Reviews: Data Mining
  and Knowledge Discovery}} (\bibinfo{year}{2021}), \bibinfo{pages}{e1415}.
\newblock


\bibitem[\protect\citeauthoryear{Chawla, Bowyer, Hall, and Kegelmeyer}{Chawla
  et~al\mbox{.}}{2002}]%
        {chawla2002smote}
\bibfield{author}{\bibinfo{person}{Nitesh~V Chawla}, \bibinfo{person}{Kevin~W
  Bowyer}, \bibinfo{person}{Lawrence~O Hall}, {and} \bibinfo{person}{W~Philip
  Kegelmeyer}.} \bibinfo{year}{2002}\natexlab{}.
\newblock \showarticletitle{SMOTE: synthetic minority over-sampling technique}.
\newblock \bibinfo{journal}{\emph{Journal of artificial intelligence research}}
   \bibinfo{volume}{16} (\bibinfo{year}{2002}), \bibinfo{pages}{321--357}.
\newblock


\bibitem[\protect\citeauthoryear{Choi and Lee}{Choi and Lee}{2019}]%
        {choi2019embracenet}
\bibfield{author}{\bibinfo{person}{Jun-Ho Choi} {and}
  \bibinfo{person}{Jong-Seok Lee}.} \bibinfo{year}{2019}\natexlab{}.
\newblock \showarticletitle{EmbraceNet: A robust deep learning architecture for
  multimodal classification}.
\newblock \bibinfo{journal}{\emph{Information Fusion}}  \bibinfo{volume}{51}
  (\bibinfo{year}{2019}), \bibinfo{pages}{259--270}.
\newblock


\bibitem[\protect\citeauthoryear{D{\"a}hne, Biessmann, Samek, Haufe, Goltz,
  Gundlach, Villringer, Fazli, and M{\"u}ller}{D{\"a}hne et~al\mbox{.}}{2015}]%
        {dahne2015multivariate}
\bibfield{author}{\bibinfo{person}{Sven D{\"a}hne}, \bibinfo{person}{Felix
  Biessmann}, \bibinfo{person}{Wojciech Samek}, \bibinfo{person}{Stefan Haufe},
  \bibinfo{person}{Dominique Goltz}, \bibinfo{person}{Christopher Gundlach},
  \bibinfo{person}{Arno Villringer}, \bibinfo{person}{Siamac Fazli}, {and}
  \bibinfo{person}{Klaus-Robert M{\"u}ller}.} \bibinfo{year}{2015}\natexlab{}.
\newblock \showarticletitle{Multivariate machine learning methods for fusing
  multimodal functional neuroimaging data}.
\newblock \bibinfo{journal}{\emph{Proc. IEEE}} \bibinfo{volume}{103},
  \bibinfo{number}{9} (\bibinfo{year}{2015}), \bibinfo{pages}{1507--1530}.
\newblock


\bibitem[\protect\citeauthoryear{Di~Mitri, Schneider, Specht, and
  Drachsler}{Di~Mitri et~al\mbox{.}}{2018}]%
        {di2018signals}
\bibfield{author}{\bibinfo{person}{Daniele Di~Mitri}, \bibinfo{person}{Jan
  Schneider}, \bibinfo{person}{Marcus Specht}, {and} \bibinfo{person}{Hendrik
  Drachsler}.} \bibinfo{year}{2018}\natexlab{}.
\newblock \showarticletitle{From signals to knowledge: A conceptual model for
  multimodal learning analytics}.
\newblock \bibinfo{journal}{\emph{Journal of Computer Assisted Learning}}
  \bibinfo{volume}{34}, \bibinfo{number}{4} (\bibinfo{year}{2018}),
  \bibinfo{pages}{338--349}.
\newblock


\bibitem[\protect\citeauthoryear{Elola, Aramendi, Irusta, Berve, and Wik}{Elola
  et~al\mbox{.}}{2020}]%
        {elola2020multimodal}
\bibfield{author}{\bibinfo{person}{Andoni Elola}, \bibinfo{person}{Elisabete
  Aramendi}, \bibinfo{person}{Unai Irusta}, \bibinfo{person}{Per~Olav Berve},
  {and} \bibinfo{person}{Lars Wik}.} \bibinfo{year}{2020}\natexlab{}.
\newblock \showarticletitle{Multimodal algorithms for the classification of
  circulation states during out-of-hospital cardiac arrest}.
\newblock \bibinfo{journal}{\emph{IEEE Transactions on Biomedical Engineering}}
  \bibinfo{volume}{68}, \bibinfo{number}{6} (\bibinfo{year}{2020}),
  \bibinfo{pages}{1913--1922}.
\newblock


\bibitem[\protect\citeauthoryear{Erickson, Xing, Srirangam, Chernova, and
  Kemp}{Erickson et~al\mbox{.}}{2020}]%
        {erickson2020multimodal}
\bibfield{author}{\bibinfo{person}{Zackory Erickson}, \bibinfo{person}{Eliot
  Xing}, \bibinfo{person}{Bharat Srirangam}, \bibinfo{person}{Sonia Chernova},
  {and} \bibinfo{person}{Charles~C Kemp}.} \bibinfo{year}{2020}\natexlab{}.
\newblock \showarticletitle{Multimodal material classification for robots using
  spectroscopy and high resolution texture imaging}. In
  \bibinfo{booktitle}{\emph{2020 IEEE/RSJ International Conference on
  Intelligent Robots and Systems (IROS)}}. IEEE, \bibinfo{pages}{10452--10459}.
\newblock


\bibitem[\protect\citeauthoryear{Felipe, Zanoni, Sehaber-Sierakowski,
  Bossolani, Souza, Flores, Oliveira, Pereira, and Costa}{Felipe
  et~al\mbox{.}}{2021}]%
        {felipe2021automatic}
\bibfield{author}{\bibinfo{person}{Gustavo~Z Felipe},
  \bibinfo{person}{Jacqueline~N Zanoni}, \bibinfo{person}{Camila~C
  Sehaber-Sierakowski}, \bibinfo{person}{Gleison~DP Bossolani},
  \bibinfo{person}{Sara~RG Souza}, \bibinfo{person}{Franklin~C Flores},
  \bibinfo{person}{Luiz~ES Oliveira}, \bibinfo{person}{Rodolfo~M Pereira},
  {and} \bibinfo{person}{Yandre~MG Costa}.} \bibinfo{year}{2021}\natexlab{}.
\newblock \showarticletitle{Automatic chronic degenerative diseases
  identification using enteric nervous system images}.
\newblock \bibinfo{journal}{\emph{Neural Computing and Applications}}
  (\bibinfo{year}{2021}), \bibinfo{pages}{1--23}.
\newblock


\bibitem[\protect\citeauthoryear{Gadiraju, Ramachandra, Chen, and
  Vatsavai}{Gadiraju et~al\mbox{.}}{2020}]%
        {gadiraju2020multimodal}
\bibfield{author}{\bibinfo{person}{Krishna~Karthik Gadiraju},
  \bibinfo{person}{Bharathkumar Ramachandra}, \bibinfo{person}{Zexi Chen},
  {and} \bibinfo{person}{Ranga~Raju Vatsavai}.}
  \bibinfo{year}{2020}\natexlab{}.
\newblock \showarticletitle{Multimodal Deep Learning Based Crop Classification
  Using Multispectral and Multitemporal Satellite Imagery}. In
  \bibinfo{booktitle}{\emph{Proceedings of the 26th ACM SIGKDD International
  Conference on Knowledge Discovery \& Data Mining}}.
  \bibinfo{pages}{3234--3242}.
\newblock


\bibitem[\protect\citeauthoryear{Gallo, Calefati, and Nawaz}{Gallo
  et~al\mbox{.}}{2017}]%
        {gallo2017multimodal}
\bibfield{author}{\bibinfo{person}{Ignazio Gallo}, \bibinfo{person}{Alessandro
  Calefati}, {and} \bibinfo{person}{Shah Nawaz}.}
  \bibinfo{year}{2017}\natexlab{}.
\newblock \showarticletitle{Multimodal classification fusion in real-world
  scenarios}. In \bibinfo{booktitle}{\emph{2017 14th IAPR International
  Conference on Document Analysis and Recognition (ICDAR)}},
  Vol.~\bibinfo{volume}{5}. IEEE, \bibinfo{pages}{36--41}.
\newblock


\bibitem[\protect\citeauthoryear{Gao, Li, Chen, and Zhang}{Gao
  et~al\mbox{.}}{2020}]%
        {gao2020survey}
\bibfield{author}{\bibinfo{person}{Jing Gao}, \bibinfo{person}{Peng Li},
  \bibinfo{person}{Zhikui Chen}, {and} \bibinfo{person}{Jianing Zhang}.}
  \bibinfo{year}{2020}\natexlab{}.
\newblock \showarticletitle{A survey on deep learning for multimodal data
  fusion}.
\newblock \bibinfo{journal}{\emph{Neural Computation}} \bibinfo{volume}{32},
  \bibinfo{number}{5} (\bibinfo{year}{2020}), \bibinfo{pages}{829--864}.
\newblock


\bibitem[\protect\citeauthoryear{Gao, Shi, Shen, and Liu}{Gao
  et~al\mbox{.}}{2021}]%
        {gao2021task}
\bibfield{author}{\bibinfo{person}{Xingyu Gao}, \bibinfo{person}{Feng Shi},
  \bibinfo{person}{Dinggang Shen}, {and} \bibinfo{person}{Manhua Liu}.}
  \bibinfo{year}{2021}\natexlab{}.
\newblock \showarticletitle{Task-induced Pyramid and Attention GAN for
  Multimodal Brain Image Imputation and Classification in Alzheimers disease}.
\newblock \bibinfo{journal}{\emph{IEEE Journal of Biomedical and Health
  Informatics}} (\bibinfo{year}{2021}).
\newblock


\bibitem[\protect\citeauthoryear{Garcia-Ceja, Riegler, Nordgreen, Jakobsen,
  Oedegaard, and T{\o}rresen}{Garcia-Ceja et~al\mbox{.}}{2018}]%
        {garcia2018mental}
\bibfield{author}{\bibinfo{person}{Enrique Garcia-Ceja},
  \bibinfo{person}{Michael Riegler}, \bibinfo{person}{Tine Nordgreen},
  \bibinfo{person}{Petter Jakobsen}, \bibinfo{person}{Ketil~J Oedegaard}, {and}
  \bibinfo{person}{Jim T{\o}rresen}.} \bibinfo{year}{2018}\natexlab{}.
\newblock \showarticletitle{Mental health monitoring with multimodal sensing
  and machine learning: A survey}.
\newblock \bibinfo{journal}{\emph{Pervasive and Mobile Computing}}
  \bibinfo{volume}{51} (\bibinfo{year}{2018}), \bibinfo{pages}{1--26}.
\newblock


\bibitem[\protect\citeauthoryear{Garillos-Manliguez and
  Chiang}{Garillos-Manliguez and Chiang}{2021}]%
        {garillos2021multimodal}
\bibfield{author}{\bibinfo{person}{Cinmayii~A Garillos-Manliguez} {and}
  \bibinfo{person}{John~Y Chiang}.} \bibinfo{year}{2021}\natexlab{}.
\newblock \showarticletitle{Multimodal Deep Learning and Visible-Light and
  Hyperspectral Imaging for Fruit Maturity Estimation}.
\newblock \bibinfo{journal}{\emph{Sensors}} \bibinfo{volume}{21},
  \bibinfo{number}{4} (\bibinfo{year}{2021}), \bibinfo{pages}{1288}.
\newblock


\bibitem[\protect\citeauthoryear{Griffiths and Boehm}{Griffiths and
  Boehm}{2019}]%
        {griffiths2019review}
\bibfield{author}{\bibinfo{person}{David Griffiths} {and} \bibinfo{person}{Jan
  Boehm}.} \bibinfo{year}{2019}\natexlab{}.
\newblock \showarticletitle{A review on deep learning techniques for 3D sensed
  data classification}.
\newblock \bibinfo{journal}{\emph{Remote Sensing}} \bibinfo{volume}{11},
  \bibinfo{number}{12} (\bibinfo{year}{2019}), \bibinfo{pages}{1499}.
\newblock


\bibitem[\protect\citeauthoryear{Gu, Chanussot, Jia, and Benediktsson}{Gu
  et~al\mbox{.}}{2017}]%
        {gu2017multiple}
\bibfield{author}{\bibinfo{person}{Yanfeng Gu}, \bibinfo{person}{Jocelyn
  Chanussot}, \bibinfo{person}{Xiuping Jia}, {and} \bibinfo{person}{Jon~Atli
  Benediktsson}.} \bibinfo{year}{2017}\natexlab{}.
\newblock \showarticletitle{Multiple kernel learning for hyperspectral image
  classification: A review}.
\newblock \bibinfo{journal}{\emph{IEEE Transactions on Geoscience and Remote
  Sensing}} \bibinfo{volume}{55}, \bibinfo{number}{11} (\bibinfo{year}{2017}),
  \bibinfo{pages}{6547--6565}.
\newblock


\bibitem[\protect\citeauthoryear{Guggenmos, Schmack, Veer, Lett, Sekutowicz,
  Sebold, Garbusow, Sommer, Wittchen, Zimmermann, et~al\mbox{.}}{Guggenmos
  et~al\mbox{.}}{2020}]%
        {guggenmos2020multimodal}
\bibfield{author}{\bibinfo{person}{Matthias Guggenmos},
  \bibinfo{person}{Katharina Schmack}, \bibinfo{person}{Ilya~M Veer},
  \bibinfo{person}{Tristram Lett}, \bibinfo{person}{Maria Sekutowicz},
  \bibinfo{person}{Miriam Sebold}, \bibinfo{person}{Maria Garbusow},
  \bibinfo{person}{Christian Sommer}, \bibinfo{person}{Hans-Ulrich Wittchen},
  \bibinfo{person}{Ulrich~S Zimmermann}, {et~al\mbox{.}}}
  \bibinfo{year}{2020}\natexlab{}.
\newblock \showarticletitle{A multimodal neuroimaging classifier for alcohol
  dependence}.
\newblock \bibinfo{journal}{\emph{Scientific reports}} \bibinfo{volume}{10},
  \bibinfo{number}{1} (\bibinfo{year}{2020}), \bibinfo{pages}{1--12}.
\newblock


\bibitem[\protect\citeauthoryear{Guo, Wang, and Wang}{Guo
  et~al\mbox{.}}{2019}]%
        {guo2019deep}
\bibfield{author}{\bibinfo{person}{Wenzhong Guo}, \bibinfo{person}{Jianwen
  Wang}, {and} \bibinfo{person}{Shiping Wang}.}
  \bibinfo{year}{2019}\natexlab{}.
\newblock \showarticletitle{Deep multimodal representation learning: A survey}.
\newblock \bibinfo{journal}{\emph{IEEE Access}}  \bibinfo{volume}{7}
  (\bibinfo{year}{2019}), \bibinfo{pages}{63373--63394}.
\newblock


\bibitem[\protect\citeauthoryear{Gupta, Kim, Kim, and Kwon}{Gupta
  et~al\mbox{.}}{2020}]%
        {gupta2020classification}
\bibfield{author}{\bibinfo{person}{Yubraj Gupta}, \bibinfo{person}{Ji-In Kim},
  \bibinfo{person}{Byeong~Chae Kim}, {and} \bibinfo{person}{Goo-Rak Kwon}.}
  \bibinfo{year}{2020}\natexlab{}.
\newblock \showarticletitle{Classification and graphical analysis of
  Alzheimer’s disease and its prodromal stage using multimodal features from
  structural, diffusion, and functional neuroimaging data and the APOE
  genotype}.
\newblock \bibinfo{journal}{\emph{Frontiers in aging neuroscience}}
  \bibinfo{volume}{12} (\bibinfo{year}{2020}), \bibinfo{pages}{238}.
\newblock


\bibitem[\protect\citeauthoryear{Han, Wang, and Mao}{Han et~al\mbox{.}}{2005}]%
        {borderlineSMOTE}
\bibfield{author}{\bibinfo{person}{Hui Han}, \bibinfo{person}{Wen-Yuan Wang},
  {and} \bibinfo{person}{Bing-Huan Mao}.} \bibinfo{year}{2005}\natexlab{}.
\newblock \showarticletitle{Borderline-SMOTE: A New Over-Sampling Method in
  Imbalanced Data Sets Learning}. In \bibinfo{booktitle}{\emph{Advances in
  Intelligent Computing}}, \bibfield{editor}{\bibinfo{person}{De-Shuang Huang},
  \bibinfo{person}{Xiao-Ping Zhang}, {and} \bibinfo{person}{Guang-Bin Huang}}
  (Eds.). \bibinfo{publisher}{Springer Berlin Heidelberg},
  \bibinfo{address}{Berlin, Heidelberg}, \bibinfo{pages}{878--887}.
\newblock
\showISBNx{978-3-540-31902-3}


\bibitem[\protect\citeauthoryear{Haskins, Kruger, and Yan}{Haskins
  et~al\mbox{.}}{2020}]%
        {haskins2020deep}
\bibfield{author}{\bibinfo{person}{Grant Haskins}, \bibinfo{person}{Uwe
  Kruger}, {and} \bibinfo{person}{Pingkun Yan}.}
  \bibinfo{year}{2020}\natexlab{}.
\newblock \showarticletitle{Deep learning in medical image registration: a
  survey}.
\newblock \bibinfo{journal}{\emph{Machine Vision and Applications}}
  \bibinfo{volume}{31}, \bibinfo{number}{1} (\bibinfo{year}{2020}),
  \bibinfo{pages}{1--18}.
\newblock


\bibitem[\protect\citeauthoryear{He and Garcia}{He and Garcia}{2009}]%
        {he2009learning}
\bibfield{author}{\bibinfo{person}{Haibo He} {and} \bibinfo{person}{Edwardo~A
  Garcia}.} \bibinfo{year}{2009}\natexlab{}.
\newblock \showarticletitle{Learning from imbalanced data}.
\newblock \bibinfo{journal}{\emph{IEEE Transactions on knowledge and data
  engineering}} \bibinfo{volume}{21}, \bibinfo{number}{9}
  (\bibinfo{year}{2009}), \bibinfo{pages}{1263--1284}.
\newblock


\bibitem[\protect\citeauthoryear{He, Zhang, Ren, and Sun}{He
  et~al\mbox{.}}{2016}]%
        {he2016deep}
\bibfield{author}{\bibinfo{person}{Kaiming He}, \bibinfo{person}{Xiangyu
  Zhang}, \bibinfo{person}{Shaoqing Ren}, {and} \bibinfo{person}{Jian Sun}.}
  \bibinfo{year}{2016}\natexlab{}.
\newblock \showarticletitle{Deep residual learning for image recognition}. In
  \bibinfo{booktitle}{\emph{Proceedings of the IEEE conference on computer
  vision and pattern recognition}}. \bibinfo{pages}{770--778}.
\newblock


\bibitem[\protect\citeauthoryear{Henderson, Rowe, Paquette, Baker, and
  Lester}{Henderson et~al\mbox{.}}{2020}]%
        {henderson2020improving}
\bibfield{author}{\bibinfo{person}{Nathan Henderson}, \bibinfo{person}{Jonathan
  Rowe}, \bibinfo{person}{Luc Paquette}, \bibinfo{person}{Ryan~S Baker}, {and}
  \bibinfo{person}{James Lester}.} \bibinfo{year}{2020}\natexlab{}.
\newblock \showarticletitle{Improving affect detection in game-based learning
  with multimodal data fusion}. In \bibinfo{booktitle}{\emph{International
  Conference on Artificial Intelligence in Education}}. Springer,
  \bibinfo{pages}{228--239}.
\newblock


\bibitem[\protect\citeauthoryear{Hong, Gao, Yokoya, Yao, Chanussot, Du, and
  Zhang}{Hong et~al\mbox{.}}{2020}]%
        {hong2020more}
\bibfield{author}{\bibinfo{person}{Danfeng Hong}, \bibinfo{person}{Lianru Gao},
  \bibinfo{person}{Naoto Yokoya}, \bibinfo{person}{Jing Yao},
  \bibinfo{person}{Jocelyn Chanussot}, \bibinfo{person}{Qian Du}, {and}
  \bibinfo{person}{Bing Zhang}.} \bibinfo{year}{2020}\natexlab{}.
\newblock \showarticletitle{More diverse means better: Multimodal deep learning
  meets remote-sensing imagery classification}.
\newblock \bibinfo{journal}{\emph{IEEE Transactions on Geoscience and Remote
  Sensing}} (\bibinfo{year}{2020}).
\newblock


\bibitem[\protect\citeauthoryear{Huang, Yang, Yin, Mo, and Zhong}{Huang
  et~al\mbox{.}}{2020b}]%
        {huang2020review}
\bibfield{author}{\bibinfo{person}{Bing Huang}, \bibinfo{person}{Feng Yang},
  \bibinfo{person}{Mengxiao Yin}, \bibinfo{person}{Xiaoying Mo}, {and}
  \bibinfo{person}{Cheng Zhong}.} \bibinfo{year}{2020}\natexlab{b}.
\newblock \showarticletitle{A review of multimodal medical image fusion
  techniques}.
\newblock \bibinfo{journal}{\emph{Computational and mathematical methods in
  medicine}}  \bibinfo{volume}{2020} (\bibinfo{year}{2020}).
\newblock


\bibitem[\protect\citeauthoryear{Huang, Zhou, Ding, and Zhang}{Huang
  et~al\mbox{.}}{2011}]%
        {huang2011extreme}
\bibfield{author}{\bibinfo{person}{Guang-Bin Huang}, \bibinfo{person}{Hongming
  Zhou}, \bibinfo{person}{Xiaojian Ding}, {and} \bibinfo{person}{Rui Zhang}.}
  \bibinfo{year}{2011}\natexlab{}.
\newblock \showarticletitle{Extreme learning machine for regression and
  multiclass classification}.
\newblock \bibinfo{journal}{\emph{IEEE Transactions on Systems, Man, and
  Cybernetics, Part B (Cybernetics)}} \bibinfo{volume}{42}, \bibinfo{number}{2}
  (\bibinfo{year}{2011}), \bibinfo{pages}{513--529}.
\newblock


\bibitem[\protect\citeauthoryear{Huang, Pareek, Seyyedi, Banerjee, and
  Lungren}{Huang et~al\mbox{.}}{2020a}]%
        {huang2020fusion}
\bibfield{author}{\bibinfo{person}{Shih-Cheng Huang}, \bibinfo{person}{Anuj
  Pareek}, \bibinfo{person}{Saeed Seyyedi}, \bibinfo{person}{Imon Banerjee},
  {and} \bibinfo{person}{Matthew~P Lungren}.} \bibinfo{year}{2020}\natexlab{a}.
\newblock \showarticletitle{Fusion of medical imaging and electronic health
  records using deep learning: a systematic review and implementation
  guidelines}.
\newblock \bibinfo{journal}{\emph{NPJ digital medicine}} \bibinfo{volume}{3},
  \bibinfo{number}{1} (\bibinfo{year}{2020}), \bibinfo{pages}{1--9}.
\newblock


\bibitem[\protect\citeauthoryear{Huddar, Sannakki, and Rajpurohit}{Huddar
  et~al\mbox{.}}{2018}]%
        {huddar2018ensemble}
\bibfield{author}{\bibinfo{person}{Mahesh~G Huddar}, \bibinfo{person}{Sanjeev~S
  Sannakki}, {and} \bibinfo{person}{Vijay~S Rajpurohit}.}
  \bibinfo{year}{2018}\natexlab{}.
\newblock \showarticletitle{An ensemble approach to utterance level multimodal
  sentiment analysis}. In \bibinfo{booktitle}{\emph{2018 International
  Conference on Computational Techniques, Electronics and Mechanical Systems
  (CTEMS)}}. IEEE, \bibinfo{pages}{145--150}.
\newblock


\bibitem[\protect\citeauthoryear{Huddar, Sannakki, and Rajpurohit}{Huddar
  et~al\mbox{.}}{2020}]%
        {huddar2020multi}
\bibfield{author}{\bibinfo{person}{Mahesh~G Huddar}, \bibinfo{person}{Sanjeev~S
  Sannakki}, {and} \bibinfo{person}{Vijay~S Rajpurohit}.}
  \bibinfo{year}{2020}\natexlab{}.
\newblock \showarticletitle{Multi-level feature optimization and multimodal
  contextual fusion for sentiment analysis and emotion classification}.
\newblock \bibinfo{journal}{\emph{Computational Intelligence}}
  \bibinfo{volume}{36}, \bibinfo{number}{2} (\bibinfo{year}{2020}),
  \bibinfo{pages}{861--881}.
\newblock


\bibitem[\protect\citeauthoryear{Ieracitano, Mammone, Hussain, and
  Morabito}{Ieracitano et~al\mbox{.}}{2020}]%
        {ieracitano2020novel}
\bibfield{author}{\bibinfo{person}{Cosimo Ieracitano}, \bibinfo{person}{Nadia
  Mammone}, \bibinfo{person}{Amir Hussain}, {and} \bibinfo{person}{Francesco~C
  Morabito}.} \bibinfo{year}{2020}\natexlab{}.
\newblock \showarticletitle{A novel multi-modal machine learning based approach
  for automatic classification of EEG recordings in dementia}.
\newblock \bibinfo{journal}{\emph{Neural Networks}}  \bibinfo{volume}{123}
  (\bibinfo{year}{2020}), \bibinfo{pages}{176--190}.
\newblock


\bibitem[\protect\citeauthoryear{Illendula and Sheth}{Illendula and
  Sheth}{2019}]%
        {illendula2019multimodal}
\bibfield{author}{\bibinfo{person}{Anurag Illendula} {and}
  \bibinfo{person}{Amit Sheth}.} \bibinfo{year}{2019}\natexlab{}.
\newblock \showarticletitle{Multimodal emotion classification}. In
  \bibinfo{booktitle}{\emph{Companion Proceedings of The 2019 World Wide Web
  Conference}}. \bibinfo{pages}{439--449}.
\newblock


\bibitem[\protect\citeauthoryear{Jafari, Ganesan, Thalisetty, Sivasubramanian,
  Oates, and Mohsenin}{Jafari et~al\mbox{.}}{2018}]%
        {jafari2018sensornet}
\bibfield{author}{\bibinfo{person}{Ali Jafari}, \bibinfo{person}{Ashwinkumar
  Ganesan}, \bibinfo{person}{Chetan Sai~Kumar Thalisetty},
  \bibinfo{person}{Varun Sivasubramanian}, \bibinfo{person}{Tim Oates}, {and}
  \bibinfo{person}{Tinoosh Mohsenin}.} \bibinfo{year}{2018}\natexlab{}.
\newblock \showarticletitle{Sensornet: A scalable and low-power deep
  convolutional neural network for multimodal data classification}.
\newblock \bibinfo{journal}{\emph{IEEE Transactions on Circuits and Systems I:
  Regular Papers}} \bibinfo{volume}{66}, \bibinfo{number}{1}
  (\bibinfo{year}{2018}), \bibinfo{pages}{274--287}.
\newblock


\bibitem[\protect\citeauthoryear{Jaiswal, Aldeneh, and Mower~Provost}{Jaiswal
  et~al\mbox{.}}{2019}]%
        {jaiswal2019controlling}
\bibfield{author}{\bibinfo{person}{Mimansa Jaiswal}, \bibinfo{person}{Zakaria
  Aldeneh}, {and} \bibinfo{person}{Emily Mower~Provost}.}
  \bibinfo{year}{2019}\natexlab{}.
\newblock \showarticletitle{Controlling for confounders in multimodal emotion
  classification via adversarial learning}. In \bibinfo{booktitle}{\emph{2019
  International Conference on Multimodal Interaction}}.
  \bibinfo{pages}{174--184}.
\newblock


\bibitem[\protect\citeauthoryear{Jiang, Ma, Xiao, Shao, and Guo}{Jiang
  et~al\mbox{.}}{2021}]%
        {jiang2021review}
\bibfield{author}{\bibinfo{person}{Xingyu Jiang}, \bibinfo{person}{Jiayi Ma},
  \bibinfo{person}{Guobao Xiao}, \bibinfo{person}{Zhenfeng Shao}, {and}
  \bibinfo{person}{Xiaojie Guo}.} \bibinfo{year}{2021}\natexlab{}.
\newblock \showarticletitle{A review of multimodal image matching: Methods and
  applications}.
\newblock \bibinfo{journal}{\emph{Information Fusion}} (\bibinfo{year}{2021}).
\newblock


\bibitem[\protect\citeauthoryear{Kang and Kang}{Kang and Kang}{2017}]%
        {kang2017prediction}
\bibfield{author}{\bibinfo{person}{Hyeon-Woo Kang} {and}
  \bibinfo{person}{Hang-Bong Kang}.} \bibinfo{year}{2017}\natexlab{}.
\newblock \showarticletitle{Prediction of crime occurrence from multi-modal
  data using deep learning}.
\newblock \bibinfo{journal}{\emph{PloS one}} \bibinfo{volume}{12},
  \bibinfo{number}{4} (\bibinfo{year}{2017}), \bibinfo{pages}{e0176244}.
\newblock


\bibitem[\protect\citeauthoryear{Kautzky, Vanicek, Philippe, Kranz, Wadsak,
  Mitterhauser, Hartmann, Hahn, Hacker, Rujescu, et~al\mbox{.}}{Kautzky
  et~al\mbox{.}}{2020}]%
        {kautzky2020machine}
\bibfield{author}{\bibinfo{person}{A Kautzky}, \bibinfo{person}{T Vanicek},
  \bibinfo{person}{C Philippe}, \bibinfo{person}{GS Kranz}, \bibinfo{person}{W
  Wadsak}, \bibinfo{person}{M Mitterhauser}, \bibinfo{person}{A Hartmann},
  \bibinfo{person}{A Hahn}, \bibinfo{person}{M Hacker}, \bibinfo{person}{D
  Rujescu}, {et~al\mbox{.}}} \bibinfo{year}{2020}\natexlab{}.
\newblock \showarticletitle{Machine learning classification of ADHD and HC by
  multimodal serotonergic data}.
\newblock \bibinfo{journal}{\emph{Translational psychiatry}}
  \bibinfo{volume}{10}, \bibinfo{number}{1} (\bibinfo{year}{2020}),
  \bibinfo{pages}{1--9}.
\newblock


\bibitem[\protect\citeauthoryear{Krawczyk}{Krawczyk}{2016}]%
        {krawczyk2016learning}
\bibfield{author}{\bibinfo{person}{Bartosz Krawczyk}.}
  \bibinfo{year}{2016}\natexlab{}.
\newblock \showarticletitle{Learning from imbalanced data: open challenges and
  future directions}.
\newblock \bibinfo{journal}{\emph{Progress in Artificial Intelligence}}
  \bibinfo{volume}{5}, \bibinfo{number}{4} (\bibinfo{year}{2016}),
  \bibinfo{pages}{221--232}.
\newblock


\bibitem[\protect\citeauthoryear{Lee, Mawla, Kim, Loggia, Ortiz, Jung, Chan,
  Gerber, Schmithorst, Edwards, et~al\mbox{.}}{Lee et~al\mbox{.}}{2019}]%
        {lee2019machine}
\bibfield{author}{\bibinfo{person}{Jeungchan Lee}, \bibinfo{person}{Ishtiaq
  Mawla}, \bibinfo{person}{Jieun Kim}, \bibinfo{person}{Marco~L Loggia},
  \bibinfo{person}{Ana Ortiz}, \bibinfo{person}{Changjin Jung},
  \bibinfo{person}{Suk-Tak Chan}, \bibinfo{person}{Jessica Gerber},
  \bibinfo{person}{Vincent~J Schmithorst}, \bibinfo{person}{Robert~R Edwards},
  {et~al\mbox{.}}} \bibinfo{year}{2019}\natexlab{}.
\newblock \showarticletitle{Machine learning-based prediction of clinical pain
  using multimodal neuroimaging and autonomic metrics}.
\newblock \bibinfo{journal}{\emph{Pain}} \bibinfo{volume}{160},
  \bibinfo{number}{3} (\bibinfo{year}{2019}), \bibinfo{pages}{550}.
\newblock


\bibitem[\protect\citeauthoryear{Li, Shrestha, Fioranelli, Le~Kernec, and
  Heidari}{Li et~al\mbox{.}}{2018a}]%
        {li2018hierarchical}
\bibfield{author}{\bibinfo{person}{Haobo Li}, \bibinfo{person}{Aman Shrestha},
  \bibinfo{person}{Francesco Fioranelli}, \bibinfo{person}{Julien Le~Kernec},
  {and} \bibinfo{person}{Hadi Heidari}.} \bibinfo{year}{2018}\natexlab{a}.
\newblock \showarticletitle{Hierarchical classification on multimodal sensing
  for human activity recogintion and fall detection}. In
  \bibinfo{booktitle}{\emph{2018 IEEE SENSORS}}. IEEE, \bibinfo{pages}{1--4}.
\newblock


\bibitem[\protect\citeauthoryear{Li, Yu, Wang, and Liu}{Li
  et~al\mbox{.}}{2020b}]%
        {li2020deep}
\bibfield{author}{\bibinfo{person}{Qing Li}, \bibinfo{person}{Guanyuan Yu},
  \bibinfo{person}{Jun Wang}, {and} \bibinfo{person}{Yuehao Liu}.}
  \bibinfo{year}{2020}\natexlab{b}.
\newblock \showarticletitle{A deep multimodal generative and fusion framework
  for class-imbalanced multimodal data}.
\newblock \bibinfo{journal}{\emph{Multimedia Tools and Applications}}
  \bibinfo{volume}{79}, \bibinfo{number}{33} (\bibinfo{year}{2020}),
  \bibinfo{pages}{25023--25050}.
\newblock


\bibitem[\protect\citeauthoryear{Li, Tang, Chen, and Wang}{Li
  et~al\mbox{.}}{2019}]%
        {li2019multimodal}
\bibfield{author}{\bibinfo{person}{Xianju Li}, \bibinfo{person}{Zhuang Tang},
  \bibinfo{person}{Weitao Chen}, {and} \bibinfo{person}{Lizhe Wang}.}
  \bibinfo{year}{2019}\natexlab{}.
\newblock \showarticletitle{Multimodal and multi-model deep fusion for fine
  classification of regional complex landscape areas using ZiYuan-3 imagery}.
\newblock \bibinfo{journal}{\emph{Remote Sensing}} \bibinfo{volume}{11},
  \bibinfo{number}{22} (\bibinfo{year}{2019}), \bibinfo{pages}{2716}.
\newblock


\bibitem[\protect\citeauthoryear{Li, Wu, and Ngom}{Li et~al\mbox{.}}{2018b}]%
        {li2018review}
\bibfield{author}{\bibinfo{person}{Yifeng Li}, \bibinfo{person}{Fang-Xiang Wu},
  {and} \bibinfo{person}{Alioune Ngom}.} \bibinfo{year}{2018}\natexlab{b}.
\newblock \showarticletitle{A review on machine learning principles for
  multi-view biological data integration}.
\newblock \bibinfo{journal}{\emph{Briefings in bioinformatics}}
  \bibinfo{volume}{19}, \bibinfo{number}{2} (\bibinfo{year}{2018}),
  \bibinfo{pages}{325--340}.
\newblock


\bibitem[\protect\citeauthoryear{Li, Yang, and Zhang}{Li
  et~al\mbox{.}}{2018c}]%
        {li2018survey}
\bibfield{author}{\bibinfo{person}{Yingming Li}, \bibinfo{person}{Ming Yang},
  {and} \bibinfo{person}{Zhongfei Zhang}.} \bibinfo{year}{2018}\natexlab{c}.
\newblock \showarticletitle{A survey of multi-view representation learning}.
\newblock \bibinfo{journal}{\emph{IEEE transactions on knowledge and data
  engineering}} \bibinfo{volume}{31}, \bibinfo{number}{10}
  (\bibinfo{year}{2018}), \bibinfo{pages}{1863--1883}.
\newblock


\bibitem[\protect\citeauthoryear{Li, Wang, Li, Liu, Ye, Song, Yuan, Yuan, Xia,
  Zhang, et~al\mbox{.}}{Li et~al\mbox{.}}{2020a}]%
        {li2020multi}
\bibfield{author}{\bibinfo{person}{Zheng-Yan Li}, \bibinfo{person}{Xiao-Dong
  Wang}, \bibinfo{person}{Mou Li}, \bibinfo{person}{Xi-Jiao Liu},
  \bibinfo{person}{Zheng Ye}, \bibinfo{person}{Bin Song}, \bibinfo{person}{Fang
  Yuan}, \bibinfo{person}{Yuan Yuan}, \bibinfo{person}{Chun-Chao Xia},
  \bibinfo{person}{Xin Zhang}, {et~al\mbox{.}}}
  \bibinfo{year}{2020}\natexlab{a}.
\newblock \showarticletitle{Multi-modal radiomics model to predict treatment
  response to neoadjuvant chemotherapy for locally advanced rectal cancer}.
\newblock \bibinfo{journal}{\emph{World journal of gastroenterology}}
  \bibinfo{volume}{26}, \bibinfo{number}{19} (\bibinfo{year}{2020}),
  \bibinfo{pages}{2388}.
\newblock


\bibitem[\protect\citeauthoryear{Liang, Li, Zhang, Kong, Wang, Deng, Li, Zhao,
  Li, Meng, et~al\mbox{.}}{Liang et~al\mbox{.}}{2019}]%
        {liang2019classification}
\bibfield{author}{\bibinfo{person}{Sugai Liang}, \bibinfo{person}{Yinfei Li},
  \bibinfo{person}{Zhong Zhang}, \bibinfo{person}{Xiangzhen Kong},
  \bibinfo{person}{Qiang Wang}, \bibinfo{person}{Wei Deng},
  \bibinfo{person}{Xiaojing Li}, \bibinfo{person}{Liansheng Zhao},
  \bibinfo{person}{Mingli Li}, \bibinfo{person}{Yajing Meng}, {et~al\mbox{.}}}
  \bibinfo{year}{2019}\natexlab{}.
\newblock \showarticletitle{Classification of first-episode schizophrenia using
  multimodal brain features: a combined structural and diffusion imaging
  study}.
\newblock \bibinfo{journal}{\emph{Schizophrenia bulletin}}
  \bibinfo{volume}{45}, \bibinfo{number}{3} (\bibinfo{year}{2019}),
  \bibinfo{pages}{591--599}.
\newblock


\bibitem[\protect\citeauthoryear{Liang, Li, Guo, Yu, Zheng, Samtani, and
  Zeng}{Liang et~al\mbox{.}}{2021}]%
        {liang2021fusion}
\bibfield{author}{\bibinfo{person}{Yunji Liang}, \bibinfo{person}{Huihui Li},
  \bibinfo{person}{Bin Guo}, \bibinfo{person}{Zhiwen Yu},
  \bibinfo{person}{Xiaolong Zheng}, \bibinfo{person}{Sagar Samtani}, {and}
  \bibinfo{person}{Daniel~D Zeng}.} \bibinfo{year}{2021}\natexlab{}.
\newblock \showarticletitle{Fusion of heterogeneous attention mechanisms in
  multi-view convolutional neural network for text classification}.
\newblock \bibinfo{journal}{\emph{Information Sciences}}  \bibinfo{volume}{548}
  (\bibinfo{year}{2021}), \bibinfo{pages}{295--312}.
\newblock


\bibitem[\protect\citeauthoryear{Lin, Gao, Yuan, Chen, Feng, Chen, Du, and
  Tong}{Lin et~al\mbox{.}}{2020}]%
        {lin2020predicting}
\bibfield{author}{\bibinfo{person}{Weiming Lin}, \bibinfo{person}{Qinquan Gao},
  \bibinfo{person}{Jiangnan Yuan}, \bibinfo{person}{Zhiying Chen},
  \bibinfo{person}{Chenwei Feng}, \bibinfo{person}{Weisheng Chen},
  \bibinfo{person}{Min Du}, {and} \bibinfo{person}{Tong Tong}.}
  \bibinfo{year}{2020}\natexlab{}.
\newblock \showarticletitle{Predicting Alzheimer’s disease conversion from
  mild cognitive impairment using an extreme learning machine-based grading
  method with multimodal data}.
\newblock \bibinfo{journal}{\emph{Frontiers in aging neuroscience}}
  \bibinfo{volume}{12} (\bibinfo{year}{2020}), \bibinfo{pages}{77}.
\newblock


\bibitem[\protect\citeauthoryear{Liu, Duan, Zhang, and Cao}{Liu
  et~al\mbox{.}}{2019}]%
        {liu2019hierarchical}
\bibfield{author}{\bibinfo{person}{Shuang Liu}, \bibinfo{person}{Linlin Duan},
  \bibinfo{person}{Zhong Zhang}, {and} \bibinfo{person}{Xiaozhong Cao}.}
  \bibinfo{year}{2019}\natexlab{}.
\newblock \showarticletitle{Hierarchical multimodal fusion for ground-based
  cloud classification in weather station networks}.
\newblock \bibinfo{journal}{\emph{IEEE Access}}  \bibinfo{volume}{7}
  (\bibinfo{year}{2019}), \bibinfo{pages}{85688--85695}.
\newblock


\bibitem[\protect\citeauthoryear{Liu and Li}{Liu and Li}{2018}]%
        {liu2018multimodal}
\bibfield{author}{\bibinfo{person}{Shuang Liu} {and} \bibinfo{person}{Mei Li}.}
  \bibinfo{year}{2018}\natexlab{}.
\newblock \showarticletitle{Multimodal GAN for energy efficiency and cloud
  classification in Internet of Things}.
\newblock \bibinfo{journal}{\emph{IEEE Internet of Things Journal}}
  \bibinfo{volume}{6}, \bibinfo{number}{4} (\bibinfo{year}{2018}),
  \bibinfo{pages}{6034--6041}.
\newblock


\bibitem[\protect\citeauthoryear{Liu, Li, Zhang, Xiao, and Cao}{Liu
  et~al\mbox{.}}{2018b}]%
        {liu2018Amultimodal}
\bibfield{author}{\bibinfo{person}{Shuang Liu}, \bibinfo{person}{Mei Li},
  \bibinfo{person}{Zhong Zhang}, \bibinfo{person}{Baihua Xiao}, {and}
  \bibinfo{person}{Xiaozhong Cao}.} \bibinfo{year}{2018}\natexlab{b}.
\newblock \showarticletitle{Multimodal ground-based cloud classification using
  joint fusion convolutional neural network}.
\newblock \bibinfo{journal}{\emph{Remote Sensing}} \bibinfo{volume}{10},
  \bibinfo{number}{6} (\bibinfo{year}{2018}), \bibinfo{pages}{822}.
\newblock


\bibitem[\protect\citeauthoryear{Liu, Huang, Liao, Pu, Liu, and Peng}{Liu
  et~al\mbox{.}}{2021}]%
        {liu2021hybrid}
\bibfield{author}{\bibinfo{person}{T Liu}, \bibinfo{person}{J Huang},
  \bibinfo{person}{T Liao}, \bibinfo{person}{R Pu}, \bibinfo{person}{S Liu},
  {and} \bibinfo{person}{Y Peng}.} \bibinfo{year}{2021}\natexlab{}.
\newblock \showarticletitle{A Hybrid Deep Learning Model for Predicting
  Molecular Subtypes of Human Breast Cancer Using Multimodal Data}.
\newblock \bibinfo{journal}{\emph{IRBM}} (\bibinfo{year}{2021}).
\newblock


\bibitem[\protect\citeauthoryear{Liu, Chen, Wu, Weidman, Lure, and Li}{Liu
  et~al\mbox{.}}{2018a}]%
        {liu2018use}
\bibfield{author}{\bibinfo{person}{Xiaonan Liu}, \bibinfo{person}{Kewei Chen},
  \bibinfo{person}{Teresa Wu}, \bibinfo{person}{David Weidman},
  \bibinfo{person}{Fleming Lure}, {and} \bibinfo{person}{Jing Li}.}
  \bibinfo{year}{2018}\natexlab{a}.
\newblock \showarticletitle{Use of multimodality imaging and artificial
  intelligence for diagnosis and prognosis of early stages of Alzheimer's
  disease}.
\newblock \bibinfo{journal}{\emph{Translational Research}}
  \bibinfo{volume}{194} (\bibinfo{year}{2018}), \bibinfo{pages}{56--67}.
\newblock


\bibitem[\protect\citeauthoryear{Ma and Jia}{Ma and Jia}{2019}]%
        {ma2019brain}
\bibfield{author}{\bibinfo{person}{Xiao Ma} {and} \bibinfo{person}{Fucang
  Jia}.} \bibinfo{year}{2019}\natexlab{}.
\newblock \showarticletitle{Brain tumor classification with multimodal MR and
  pathology images}. In \bibinfo{booktitle}{\emph{International MICCAI
  Brainlesion Workshop}}. Springer, \bibinfo{pages}{343--352}.
\newblock


\bibitem[\protect\citeauthoryear{Oloyede and Hancke}{Oloyede and
  Hancke}{2016}]%
        {oloyede2016unimodal}
\bibfield{author}{\bibinfo{person}{Muhtahir~O Oloyede} {and}
  \bibinfo{person}{Gerhard~P Hancke}.} \bibinfo{year}{2016}\natexlab{}.
\newblock \showarticletitle{Unimodal and multimodal biometric sensing systems:
  a review}.
\newblock \bibinfo{journal}{\emph{IEEE access}}  \bibinfo{volume}{4}
  (\bibinfo{year}{2016}), \bibinfo{pages}{7532--7555}.
\newblock


\bibitem[\protect\citeauthoryear{Oramas, Barbieri, Nieto~Caballero, and
  Serra}{Oramas et~al\mbox{.}}{2018}]%
        {oramas2018multimodal}
\bibfield{author}{\bibinfo{person}{Sergio Oramas}, \bibinfo{person}{Francesco
  Barbieri}, \bibinfo{person}{Oriol Nieto~Caballero}, {and}
  \bibinfo{person}{Xavier Serra}.} \bibinfo{year}{2018}\natexlab{}.
\newblock \showarticletitle{Multimodal deep learning for music genre
  classification}.
\newblock \bibinfo{journal}{\emph{Transactions of the International Society for
  Music Information Retrieval. 2018; 1 (1): 4-21.}} (\bibinfo{year}{2018}).
\newblock


\bibitem[\protect\citeauthoryear{Pahde, Puscas, Klein, and Nabi}{Pahde
  et~al\mbox{.}}{2021}]%
        {pahde2021multimodal}
\bibfield{author}{\bibinfo{person}{Frederik Pahde}, \bibinfo{person}{Mihai
  Puscas}, \bibinfo{person}{Tassilo Klein}, {and} \bibinfo{person}{Moin Nabi}.}
  \bibinfo{year}{2021}\natexlab{}.
\newblock \showarticletitle{Multimodal Prototypical Networks for Few-shot
  Learning}. In \bibinfo{booktitle}{\emph{Proceedings of the IEEE/CVF Winter
  Conference on Applications of Computer Vision}}. \bibinfo{pages}{2644--2653}.
\newblock


\bibitem[\protect\citeauthoryear{Panda, Chakladar, and Dasgupta}{Panda
  et~al\mbox{.}}{2020}]%
        {panda2020multimodal}
\bibfield{author}{\bibinfo{person}{Debadrita Panda},
  \bibinfo{person}{Debashis~Das Chakladar}, {and} \bibinfo{person}{Tanmoy
  Dasgupta}.} \bibinfo{year}{2020}\natexlab{}.
\newblock \showarticletitle{Multimodal system for emotion recognition using EEG
  and customer review}. In \bibinfo{booktitle}{\emph{Proceedings of the Global
  AI Congress 2019}}. Springer, \bibinfo{pages}{399--410}.
\newblock


\bibitem[\protect\citeauthoryear{Pennington, Socher, and Manning}{Pennington
  et~al\mbox{.}}{2014}]%
        {pennington2014glove}
\bibfield{author}{\bibinfo{person}{Jeffrey Pennington},
  \bibinfo{person}{Richard Socher}, {and} \bibinfo{person}{Christopher~D
  Manning}.} \bibinfo{year}{2014}\natexlab{}.
\newblock \showarticletitle{Glove: Global vectors for word representation}. In
  \bibinfo{booktitle}{\emph{Proceedings of the 2014 conference on empirical
  methods in natural language processing (EMNLP)}}.
  \bibinfo{pages}{1532--1543}.
\newblock


\bibitem[\protect\citeauthoryear{Qureshi, Oh, Cho, Jo, and Lee}{Qureshi
  et~al\mbox{.}}{2017}]%
        {qureshi2017multimodal}
\bibfield{author}{\bibinfo{person}{Muhammad Naveed~Iqbal Qureshi},
  \bibinfo{person}{Jooyoung Oh}, \bibinfo{person}{Dongrae Cho},
  \bibinfo{person}{Hang~Joon Jo}, {and} \bibinfo{person}{Boreom Lee}.}
  \bibinfo{year}{2017}\natexlab{}.
\newblock \showarticletitle{Multimodal discrimination of schizophrenia using
  hybrid weighted feature concatenation of brain functional connectivity and
  anatomical features with an extreme learning machine}.
\newblock \bibinfo{journal}{\emph{Frontiers in neuroinformatics}}
  \bibinfo{volume}{11} (\bibinfo{year}{2017}), \bibinfo{pages}{59}.
\newblock


\bibitem[\protect\citeauthoryear{Radu, Tong, Bhattacharya, Lane, Mascolo,
  Marina, and Kawsar}{Radu et~al\mbox{.}}{2018}]%
        {radu2018multimodal}
\bibfield{author}{\bibinfo{person}{Valentin Radu}, \bibinfo{person}{Catherine
  Tong}, \bibinfo{person}{Sourav Bhattacharya}, \bibinfo{person}{Nicholas~D
  Lane}, \bibinfo{person}{Cecilia Mascolo}, \bibinfo{person}{Mahesh~K Marina},
  {and} \bibinfo{person}{Fahim Kawsar}.} \bibinfo{year}{2018}\natexlab{}.
\newblock \showarticletitle{Multimodal deep learning for activity and context
  recognition}.
\newblock \bibinfo{journal}{\emph{Proceedings of the ACM on Interactive,
  Mobile, Wearable and Ubiquitous Technologies}} \bibinfo{volume}{1},
  \bibinfo{number}{4} (\bibinfo{year}{2018}), \bibinfo{pages}{1--27}.
\newblock


\bibitem[\protect\citeauthoryear{Ramachandram and Taylor}{Ramachandram and
  Taylor}{2017}]%
        {ramachandram2017deep}
\bibfield{author}{\bibinfo{person}{Dhanesh Ramachandram} {and}
  \bibinfo{person}{Graham~W Taylor}.} \bibinfo{year}{2017}\natexlab{}.
\newblock \showarticletitle{Deep multimodal learning: A survey on recent
  advances and trends}.
\newblock \bibinfo{journal}{\emph{IEEE Signal Processing Magazine}}
  \bibinfo{volume}{34}, \bibinfo{number}{6} (\bibinfo{year}{2017}),
  \bibinfo{pages}{96--108}.
\newblock


\bibitem[\protect\citeauthoryear{Raman and Kumari}{Raman and Kumari}{2018}]%
        {raman2018multimodal}
\bibfield{author}{\bibinfo{person}{Vishal Raman} {and} \bibinfo{person}{Madhu
  Kumari}.} \bibinfo{year}{2018}\natexlab{}.
\newblock \showarticletitle{Multimodal Deep Learning in Semantic Image
  Segmentation: A Review}. In \bibinfo{booktitle}{\emph{Proceedings of the 2018
  International Conference on Cloud Computing and Internet of Things}}.
  \bibinfo{pages}{7--11}.
\newblock


\bibitem[\protect\citeauthoryear{Rastgoo, Nakisa, Rakotonirainy, Chandran, and
  Tjondronegoro}{Rastgoo et~al\mbox{.}}{2018}]%
        {rastgoo2018critical}
\bibfield{author}{\bibinfo{person}{Mohammad~Naim Rastgoo},
  \bibinfo{person}{Bahareh Nakisa}, \bibinfo{person}{Andry Rakotonirainy},
  \bibinfo{person}{Vinod Chandran}, {and} \bibinfo{person}{Dian
  Tjondronegoro}.} \bibinfo{year}{2018}\natexlab{}.
\newblock \showarticletitle{A critical review of proactive detection of driver
  stress levels based on multimodal measurements}.
\newblock \bibinfo{journal}{\emph{ACM Computing Surveys (CSUR)}}
  \bibinfo{volume}{51}, \bibinfo{number}{5} (\bibinfo{year}{2018}),
  \bibinfo{pages}{1--35}.
\newblock


\bibitem[\protect\citeauthoryear{Rathore, Habes, Iftikhar, Shacklett, and
  Davatzikos}{Rathore et~al\mbox{.}}{2017}]%
        {rathore2017review}
\bibfield{author}{\bibinfo{person}{Saima Rathore}, \bibinfo{person}{Mohamad
  Habes}, \bibinfo{person}{Muhammad~Aksam Iftikhar}, \bibinfo{person}{Amanda
  Shacklett}, {and} \bibinfo{person}{Christos Davatzikos}.}
  \bibinfo{year}{2017}\natexlab{}.
\newblock \showarticletitle{A review on neuroimaging-based classification
  studies and associated feature extraction methods for Alzheimer's disease and
  its prodromal stages}.
\newblock \bibinfo{journal}{\emph{NeuroImage}}  \bibinfo{volume}{155}
  (\bibinfo{year}{2017}), \bibinfo{pages}{530--548}.
\newblock


\bibitem[\protect\citeauthoryear{S{\'a}ez, Krawczyk, and Wo{\'z}niak}{S{\'a}ez
  et~al\mbox{.}}{2016}]%
        {saez2016analyzing}
\bibfield{author}{\bibinfo{person}{Jos{\'e}~A S{\'a}ez},
  \bibinfo{person}{Bartosz Krawczyk}, {and} \bibinfo{person}{Micha{\l}
  Wo{\'z}niak}.} \bibinfo{year}{2016}\natexlab{}.
\newblock \showarticletitle{Analyzing the oversampling of different classes and
  types of examples in multi-class imbalanced datasets}.
\newblock \bibinfo{journal}{\emph{Pattern Recognition}}  \bibinfo{volume}{57}
  (\bibinfo{year}{2016}), \bibinfo{pages}{164--178}.
\newblock


\bibitem[\protect\citeauthoryear{Said, Mohamed, Elfouly, Harras, and Wang}{Said
  et~al\mbox{.}}{2017}]%
        {said2017multimodal}
\bibfield{author}{\bibinfo{person}{Ahmed~Ben Said}, \bibinfo{person}{Amr
  Mohamed}, \bibinfo{person}{Tarek Elfouly}, \bibinfo{person}{Khaled Harras},
  {and} \bibinfo{person}{Z~Jane Wang}.} \bibinfo{year}{2017}\natexlab{}.
\newblock \showarticletitle{Multimodal deep learning approach for joint EEG-EMG
  data compression and classification}. In \bibinfo{booktitle}{\emph{2017 IEEE
  wireless communications and networking conference (WCNC)}}. IEEE,
  \bibinfo{pages}{1--6}.
\newblock


\bibitem[\protect\citeauthoryear{Sharma, Sikka, et~al\mbox{.}}{Sharma
  et~al\mbox{.}}{2020}]%
        {sharma2020multimodal}
\bibfield{author}{\bibinfo{person}{Nonita Sharma}, \bibinfo{person}{Geeta
  Sikka}, {et~al\mbox{.}}} \bibinfo{year}{2020}\natexlab{}.
\newblock \showarticletitle{Multimodal Sentiment Analysis of Social Media Data:
  A Review}. In \bibinfo{booktitle}{\emph{The International Conference on
  Recent Innovations in Computing}}. Springer, \bibinfo{pages}{545--561}.
\newblock


\bibitem[\protect\citeauthoryear{Shen, Salehi, Baldwin, and Qi}{Shen
  et~al\mbox{.}}{2019}]%
        {shen2019joint}
\bibfield{author}{\bibinfo{person}{Aili Shen}, \bibinfo{person}{Bahar Salehi},
  \bibinfo{person}{Timothy Baldwin}, {and} \bibinfo{person}{Jianzhong Qi}.}
  \bibinfo{year}{2019}\natexlab{}.
\newblock \showarticletitle{A joint model for multimodal document quality
  assessment}. In \bibinfo{booktitle}{\emph{2019 ACM/IEEE Joint Conference on
  Digital Libraries (JCDL)}}. IEEE, \bibinfo{pages}{107--110}.
\newblock


\bibitem[\protect\citeauthoryear{Shorten and Khoshgoftaar}{Shorten and
  Khoshgoftaar}{2019}]%
        {shorten2019survey}
\bibfield{author}{\bibinfo{person}{Connor Shorten} {and}
  \bibinfo{person}{Taghi~M Khoshgoftaar}.} \bibinfo{year}{2019}\natexlab{}.
\newblock \showarticletitle{A survey on image data augmentation for deep
  learning}.
\newblock \bibinfo{journal}{\emph{Journal of Big Data}} \bibinfo{volume}{6},
  \bibinfo{number}{1} (\bibinfo{year}{2019}), \bibinfo{pages}{1--48}.
\newblock


\bibitem[\protect\citeauthoryear{Simonetta, Ntalampiras, and
  Avanzini}{Simonetta et~al\mbox{.}}{2019}]%
        {simonetta2019multimodal}
\bibfield{author}{\bibinfo{person}{Federico Simonetta},
  \bibinfo{person}{Stavros Ntalampiras}, {and} \bibinfo{person}{Federico
  Avanzini}.} \bibinfo{year}{2019}\natexlab{}.
\newblock \showarticletitle{Multimodal music information processing and
  retrieval: Survey and future challenges}. In \bibinfo{booktitle}{\emph{2019
  International Workshop on Multilayer Music Representation and Processing
  (MMRP)}}. IEEE, \bibinfo{pages}{10--18}.
\newblock


\bibitem[\protect\citeauthoryear{Simonyan and Zisserman}{Simonyan and
  Zisserman}{2014}]%
        {simonyan2014very}
\bibfield{author}{\bibinfo{person}{Karen Simonyan} {and}
  \bibinfo{person}{Andrew Zisserman}.} \bibinfo{year}{2014}\natexlab{}.
\newblock \showarticletitle{Very deep convolutional networks for large-scale
  image recognition}.
\newblock \bibinfo{journal}{\emph{arXiv preprint arXiv:1409.1556}}
  (\bibinfo{year}{2014}).
\newblock


\bibitem[\protect\citeauthoryear{Sleeman~IV and Krawczyk}{Sleeman~IV and
  Krawczyk}{2021}]%
        {sleeman2021multi}
\bibfield{author}{\bibinfo{person}{William~C Sleeman~IV} {and}
  \bibinfo{person}{Bartosz Krawczyk}.} \bibinfo{year}{2021}\natexlab{}.
\newblock \showarticletitle{Multi-class imbalanced big data classification on
  Spark}.
\newblock \bibinfo{journal}{\emph{Knowledge-Based Systems}}
  \bibinfo{volume}{212} (\bibinfo{year}{2021}), \bibinfo{pages}{106598}.
\newblock


\bibitem[\protect\citeauthoryear{Soleymani, Garcia, Jou, Schuller, Chang, and
  Pantic}{Soleymani et~al\mbox{.}}{2017}]%
        {soleymani2017survey}
\bibfield{author}{\bibinfo{person}{Mohammad Soleymani}, \bibinfo{person}{David
  Garcia}, \bibinfo{person}{Brendan Jou}, \bibinfo{person}{Bj{\"o}rn Schuller},
  \bibinfo{person}{Shih-Fu Chang}, {and} \bibinfo{person}{Maja Pantic}.}
  \bibinfo{year}{2017}\natexlab{}.
\newblock \showarticletitle{A survey of multimodal sentiment analysis}.
\newblock \bibinfo{journal}{\emph{Image and Vision Computing}}
  \bibinfo{volume}{65} (\bibinfo{year}{2017}), \bibinfo{pages}{3--14}.
\newblock


\bibitem[\protect\citeauthoryear{Song, Zheng, Ullah, Wang, Jiang, Xu, Zou, and
  Ding}{Song et~al\mbox{.}}{2021}]%
        {song2021multiview}
\bibfield{author}{\bibinfo{person}{Jingqi Song}, \bibinfo{person}{Yuanjie
  Zheng}, \bibinfo{person}{Muhammad~Zakir Ullah}, \bibinfo{person}{Junxia
  Wang}, \bibinfo{person}{Yanyun Jiang}, \bibinfo{person}{Chenxi Xu},
  \bibinfo{person}{Zhenxing Zou}, {and} \bibinfo{person}{Guocheng Ding}.}
  \bibinfo{year}{2021}\natexlab{}.
\newblock \showarticletitle{Multiview multimodal network for breast cancer
  diagnosis in contrast-enhanced spectral mammography images}.
\newblock \bibinfo{journal}{\emph{International Journal of Computer Assisted
  Radiology and Surgery}} \bibinfo{volume}{16}, \bibinfo{number}{6}
  (\bibinfo{year}{2021}), \bibinfo{pages}{979--988}.
\newblock


\bibitem[\protect\citeauthoryear{Sorinas, Ferr{\'a}ndez, and Fernandez}{Sorinas
  et~al\mbox{.}}{2020}]%
        {sorinas2020brain}
\bibfield{author}{\bibinfo{person}{Jennifer Sorinas},
  \bibinfo{person}{Jose~Manuel Ferr{\'a}ndez}, {and} \bibinfo{person}{Eduardo
  Fernandez}.} \bibinfo{year}{2020}\natexlab{}.
\newblock \showarticletitle{Brain and body emotional responses: Multimodal
  approximation for valence classification}.
\newblock \bibinfo{journal}{\emph{Sensors}} \bibinfo{volume}{20},
  \bibinfo{number}{1} (\bibinfo{year}{2020}), \bibinfo{pages}{313}.
\newblock


\bibitem[\protect\citeauthoryear{Sun}{Sun}{2013}]%
        {sun2013survey}
\bibfield{author}{\bibinfo{person}{Shiliang Sun}.}
  \bibinfo{year}{2013}\natexlab{}.
\newblock \showarticletitle{A survey of multi-view machine learning}.
\newblock \bibinfo{journal}{\emph{Neural computing and applications}}
  \bibinfo{volume}{23}, \bibinfo{number}{7} (\bibinfo{year}{2013}),
  \bibinfo{pages}{2031--2038}.
\newblock


\bibitem[\protect\citeauthoryear{Suzuki, Rin, Maeda, and Takeda}{Suzuki
  et~al\mbox{.}}{2018}]%
        {suzuki2018forest}
\bibfield{author}{\bibinfo{person}{Kumiko Suzuki}, \bibinfo{person}{Utei Rin},
  \bibinfo{person}{Yoshiko Maeda}, {and} \bibinfo{person}{Hiroshi Takeda}.}
  \bibinfo{year}{2018}\natexlab{}.
\newblock \showarticletitle{FOREST COVER CLASSIFICATION USING GEOSPATIAL
  MULTIMODAL DATA.}
\newblock \bibinfo{journal}{\emph{International Archives of the Photogrammetry,
  Remote Sensing \& Spatial Information Sciences}} \bibinfo{volume}{42},
  \bibinfo{number}{2} (\bibinfo{year}{2018}).
\newblock


\bibitem[\protect\citeauthoryear{Syed, Sleeman, Hagan, Palta, Kapoor, and
  Ghosh}{Syed et~al\mbox{.}}{2021b}]%
        {syed2021multi}
\bibfield{author}{\bibinfo{person}{Khajamoinuddin Syed},
  \bibinfo{person}{William~C Sleeman}, \bibinfo{person}{Michael Hagan},
  \bibinfo{person}{Jatinder Palta}, \bibinfo{person}{Rishabh Kapoor}, {and}
  \bibinfo{person}{Preetam Ghosh}.} \bibinfo{year}{2021}\natexlab{b}.
\newblock \showarticletitle{Multi-View Data Integration Methods for
  Radiotherapy Structure Name Standardization}.
\newblock \bibinfo{journal}{\emph{Cancers}} \bibinfo{volume}{13},
  \bibinfo{number}{8} (\bibinfo{year}{2021}), \bibinfo{pages}{1796}.
\newblock


\bibitem[\protect\citeauthoryear{Syed, Pirogova, and Lech}{Syed
  et~al\mbox{.}}{2021a}]%
        {syed2021prediction}
\bibfield{author}{\bibinfo{person}{Muhammad Shehram~Shah Syed},
  \bibinfo{person}{Elena Pirogova}, {and} \bibinfo{person}{Margaret Lech}.}
  \bibinfo{year}{2021}\natexlab{a}.
\newblock \showarticletitle{Prediction of Public Trust in Politicians Using a
  Multimodal Fusion Approach}.
\newblock \bibinfo{journal}{\emph{Electronics}} \bibinfo{volume}{10},
  \bibinfo{number}{11} (\bibinfo{year}{2021}), \bibinfo{pages}{1259}.
\newblock


\bibitem[\protect\citeauthoryear{Szegedy, Vanhoucke, Ioffe, Shlens, and
  Wojna}{Szegedy et~al\mbox{.}}{2016}]%
        {szegedy2016rethinking}
\bibfield{author}{\bibinfo{person}{Christian Szegedy}, \bibinfo{person}{Vincent
  Vanhoucke}, \bibinfo{person}{Sergey Ioffe}, \bibinfo{person}{Jon Shlens},
  {and} \bibinfo{person}{Zbigniew Wojna}.} \bibinfo{year}{2016}\natexlab{}.
\newblock \showarticletitle{Rethinking the inception architecture for computer
  vision}. In \bibinfo{booktitle}{\emph{Proceedings of the IEEE conference on
  computer vision and pattern recognition}}. \bibinfo{pages}{2818--2826}.
\newblock


\bibitem[\protect\citeauthoryear{Takahashi, Gygli, and Van~Gool}{Takahashi
  et~al\mbox{.}}{2017}]%
        {takahashi2017aenet}
\bibfield{author}{\bibinfo{person}{Naoya Takahashi}, \bibinfo{person}{Michael
  Gygli}, {and} \bibinfo{person}{Luc Van~Gool}.}
  \bibinfo{year}{2017}\natexlab{}.
\newblock \showarticletitle{Aenet: Learning deep audio features for video
  analysis}.
\newblock \bibinfo{journal}{\emph{IEEE Transactions on Multimedia}}
  \bibinfo{volume}{20}, \bibinfo{number}{3} (\bibinfo{year}{2017}),
  \bibinfo{pages}{513--524}.
\newblock


\bibitem[\protect\citeauthoryear{Tan, Sun, Zhang, Chen, and Liu}{Tan
  et~al\mbox{.}}{2017}]%
        {tan2017multimodal}
\bibfield{author}{\bibinfo{person}{Chuanqi Tan}, \bibinfo{person}{Fuchun Sun},
  \bibinfo{person}{Wenchang Zhang}, \bibinfo{person}{Jianhua Chen}, {and}
  \bibinfo{person}{Chunfang Liu}.} \bibinfo{year}{2017}\natexlab{}.
\newblock \showarticletitle{Multimodal classification with deep
  convolutional-recurrent neural networks for electroencephalography}. In
  \bibinfo{booktitle}{\emph{International Conference on Neural Information
  Processing}}. Springer, \bibinfo{pages}{767--776}.
\newblock


\bibitem[\protect\citeauthoryear{Tian, Tao, Pouyanfar, Chen, and Shyu}{Tian
  et~al\mbox{.}}{2019}]%
        {tian2019multimodal}
\bibfield{author}{\bibinfo{person}{Haiman Tian}, \bibinfo{person}{Yudong Tao},
  \bibinfo{person}{Samira Pouyanfar}, \bibinfo{person}{Shu-Ching Chen}, {and}
  \bibinfo{person}{Mei-Ling Shyu}.} \bibinfo{year}{2019}\natexlab{}.
\newblock \showarticletitle{Multimodal deep representation learning for video
  classification}.
\newblock \bibinfo{journal}{\emph{World Wide Web}} \bibinfo{volume}{22},
  \bibinfo{number}{3} (\bibinfo{year}{2019}), \bibinfo{pages}{1325--1341}.
\newblock


\bibitem[\protect\citeauthoryear{Uddin and Canavan}{Uddin and Canavan}{2020}]%
        {uddin2020multimodal}
\bibfield{author}{\bibinfo{person}{Md~Taufeeq Uddin} {and}
  \bibinfo{person}{Shaun Canavan}.} \bibinfo{year}{2020}\natexlab{}.
\newblock \showarticletitle{Multimodal multilevel fusion for sequential
  protective behavior detection and pain estimation}. In
  \bibinfo{booktitle}{\emph{2020 15th IEEE International Conference on
  Automatic Face and Gesture Recognition (FG 2020)(FG)}}. IEEE Computer
  Society, \bibinfo{pages}{467--471}.
\newblock


\bibitem[\protect\citeauthoryear{Usman and Rajpoot}{Usman and Rajpoot}{2017}]%
        {usman2017brain}
\bibfield{author}{\bibinfo{person}{Khalid Usman} {and} \bibinfo{person}{Kashif
  Rajpoot}.} \bibinfo{year}{2017}\natexlab{}.
\newblock \showarticletitle{Brain tumor classification from multi-modality MRI
  using wavelets and machine learning}.
\newblock \bibinfo{journal}{\emph{Pattern Analysis and Applications}}
  \bibinfo{volume}{20}, \bibinfo{number}{3} (\bibinfo{year}{2017}),
  \bibinfo{pages}{871--881}.
\newblock


\bibitem[\protect\citeauthoryear{Venugopalan, Tong, Hassanzadeh, and
  Wang}{Venugopalan et~al\mbox{.}}{2021}]%
        {venugopalan2021multimodal}
\bibfield{author}{\bibinfo{person}{Janani Venugopalan}, \bibinfo{person}{Li
  Tong}, \bibinfo{person}{Hamid~Reza Hassanzadeh}, {and} \bibinfo{person}{May~D
  Wang}.} \bibinfo{year}{2021}\natexlab{}.
\newblock \showarticletitle{Multimodal deep learning models for early detection
  of Alzheimer’s disease stage}.
\newblock \bibinfo{journal}{\emph{Scientific reports}} \bibinfo{volume}{11},
  \bibinfo{number}{1} (\bibinfo{year}{2021}), \bibinfo{pages}{1--13}.
\newblock


\bibitem[\protect\citeauthoryear{Vielzeuf, Pateux, and Jurie}{Vielzeuf
  et~al\mbox{.}}{2017}]%
        {vielzeuf2017temporal}
\bibfield{author}{\bibinfo{person}{Valentin Vielzeuf},
  \bibinfo{person}{St{\'e}phane Pateux}, {and}
  \bibinfo{person}{Fr{\'e}d{\'e}ric Jurie}.} \bibinfo{year}{2017}\natexlab{}.
\newblock \showarticletitle{Temporal multimodal fusion for video emotion
  classification in the wild}. In \bibinfo{booktitle}{\emph{Proceedings of the
  19th ACM International Conference on Multimodal Interaction}}.
  \bibinfo{pages}{569--576}.
\newblock


\bibitem[\protect\citeauthoryear{Vijay and Indumathi}{Vijay and
  Indumathi}{2018}]%
        {vijay2018multimodal}
\bibfield{author}{\bibinfo{person}{M Vijay} {and} \bibinfo{person}{G
  Indumathi}.} \bibinfo{year}{2018}\natexlab{}.
\newblock \showarticletitle{Multimodal Biometric System Using Ear and Palm Vein
  Recognition Based on GwPeSOA: Multi-SVNN for Security Applications}. In
  \bibinfo{booktitle}{\emph{International Conference On Computational Vision
  and Bio Inspired Computing}}. Springer, \bibinfo{pages}{215--231}.
\newblock


\bibitem[\protect\citeauthoryear{Vijay and Indumathi}{Vijay and
  Indumathi}{2021}]%
        {vijay2021deep}
\bibfield{author}{\bibinfo{person}{M Vijay} {and} \bibinfo{person}{G
  Indumathi}.} \bibinfo{year}{2021}\natexlab{}.
\newblock \showarticletitle{Deep belief network-based hybrid model for
  multimodal biometric system for futuristic security applications}.
\newblock \bibinfo{journal}{\emph{Journal of Information Security and
  Applications}}  \bibinfo{volume}{58} (\bibinfo{year}{2021}),
  \bibinfo{pages}{102707}.
\newblock


\bibitem[\protect\citeauthoryear{Wang, Bansal, and Frahm}{Wang
  et~al\mbox{.}}{2018}]%
        {wang2018retweet}
\bibfield{author}{\bibinfo{person}{Ke Wang}, \bibinfo{person}{Mohit Bansal},
  {and} \bibinfo{person}{Jan-Michael Frahm}.} \bibinfo{year}{2018}\natexlab{}.
\newblock \showarticletitle{Retweet wars: Tweet popularity prediction via
  dynamic multimodal regression}. In \bibinfo{booktitle}{\emph{2018 IEEE Winter
  Conference on Applications of Computer Vision (WACV)}}. IEEE,
  \bibinfo{pages}{1842--1851}.
\newblock


\bibitem[\protect\citeauthoryear{Wang, Hao, and Ji}{Wang et~al\mbox{.}}{2019}]%
        {wang2019knowledge}
\bibfield{author}{\bibinfo{person}{Shangfei Wang}, \bibinfo{person}{Longfei
  Hao}, {and} \bibinfo{person}{Qiang Ji}.} \bibinfo{year}{2019}\natexlab{}.
\newblock \showarticletitle{Knowledge-augmented multimodal deep regression
  bayesian networks for emotion video tagging}.
\newblock \bibinfo{journal}{\emph{IEEE Transactions on Multimedia}}
  \bibinfo{volume}{22}, \bibinfo{number}{4} (\bibinfo{year}{2019}),
  \bibinfo{pages}{1084--1097}.
\newblock


\bibitem[\protect\citeauthoryear{Xu, Tao, and Xu}{Xu et~al\mbox{.}}{2013}]%
        {xu2013survey}
\bibfield{author}{\bibinfo{person}{Chang Xu}, \bibinfo{person}{Dacheng Tao},
  {and} \bibinfo{person}{Chao Xu}.} \bibinfo{year}{2013}\natexlab{}.
\newblock \showarticletitle{A survey on multi-view learning}.
\newblock \bibinfo{journal}{\emph{arXiv preprint arXiv:1304.5634}}
  (\bibinfo{year}{2013}).
\newblock


\bibitem[\protect\citeauthoryear{Xu, Lam, Pang, Gao, Band, Mathur, Papay,
  Khanna, Cywinski, Maheshwari, et~al\mbox{.}}{Xu et~al\mbox{.}}{2019a}]%
        {xu2019multimodal}
\bibfield{author}{\bibinfo{person}{Keyang Xu}, \bibinfo{person}{Mike Lam},
  \bibinfo{person}{Jingzhi Pang}, \bibinfo{person}{Xin Gao},
  \bibinfo{person}{Charlotte Band}, \bibinfo{person}{Piyush Mathur},
  \bibinfo{person}{Frank Papay}, \bibinfo{person}{Ashish~K Khanna},
  \bibinfo{person}{Jacek~B Cywinski}, \bibinfo{person}{Kamal Maheshwari},
  {et~al\mbox{.}}} \bibinfo{year}{2019}\natexlab{a}.
\newblock \showarticletitle{Multimodal machine learning for automated ICD
  coding}. In \bibinfo{booktitle}{\emph{Machine Learning for Healthcare
  Conference}}. PMLR, \bibinfo{pages}{197--215}.
\newblock


\bibitem[\protect\citeauthoryear{Xu, Mao, and Chen}{Xu et~al\mbox{.}}{2019b}]%
        {xu2019multi}
\bibfield{author}{\bibinfo{person}{Nan Xu}, \bibinfo{person}{Wenji Mao}, {and}
  \bibinfo{person}{Guandan Chen}.} \bibinfo{year}{2019}\natexlab{b}.
\newblock \showarticletitle{Multi-interactive memory network for aspect based
  multimodal sentiment analysis}. In \bibinfo{booktitle}{\emph{Proceedings of
  the AAAI Conference on Artificial Intelligence}}, Vol.~\bibinfo{volume}{33}.
  \bibinfo{pages}{371--378}.
\newblock


\bibitem[\protect\citeauthoryear{Xu, Bai, Liu, Zhao, and Sun}{Xu
  et~al\mbox{.}}{2021}]%
        {xu2021mm}
\bibfield{author}{\bibinfo{person}{Xiujuan Xu}, \bibinfo{person}{Yulin Bai},
  \bibinfo{person}{Yu Liu}, \bibinfo{person}{Xiaowei Zhao}, {and}
  \bibinfo{person}{Yuzhi Sun}.} \bibinfo{year}{2021}\natexlab{}.
\newblock \showarticletitle{MM-UrbanFAC: Urban Functional Area Classification
  Model Based on Multimodal Machine Learning}.
\newblock \bibinfo{journal}{\emph{IEEE Transactions on Intelligent
  Transportation Systems}} (\bibinfo{year}{2021}).
\newblock


\bibitem[\protect\citeauthoryear{Yaman, Irem~Eyiokur, and Kemal~Ekenel}{Yaman
  et~al\mbox{.}}{2019}]%
        {yaman2019multimodal}
\bibfield{author}{\bibinfo{person}{Dogucan Yaman}, \bibinfo{person}{Fevziye
  Irem~Eyiokur}, {and} \bibinfo{person}{Hazim Kemal~Ekenel}.}
  \bibinfo{year}{2019}\natexlab{}.
\newblock \showarticletitle{Multimodal age and gender classification using ear
  and profile face images}. In \bibinfo{booktitle}{\emph{Proceedings of the
  IEEE/CVF Conference on Computer Vision and Pattern Recognition Workshops}}.
  \bibinfo{pages}{0--0}.
\newblock


\bibitem[\protect\citeauthoryear{Yan, Hu, Mao, Ye, and Yu}{Yan
  et~al\mbox{.}}{2021}]%
        {yan2021deep}
\bibfield{author}{\bibinfo{person}{Xiaoqiang Yan}, \bibinfo{person}{Shizhe Hu},
  \bibinfo{person}{Yiqiao Mao}, \bibinfo{person}{Yangdong Ye}, {and}
  \bibinfo{person}{Hui Yu}.} \bibinfo{year}{2021}\natexlab{}.
\newblock \showarticletitle{Deep multi-view learning methods: a review}.
\newblock \bibinfo{journal}{\emph{Neurocomputing}} (\bibinfo{year}{2021}).
\newblock


\bibitem[\protect\citeauthoryear{Yang, Feng, Wang, and Zhang}{Yang
  et~al\mbox{.}}{2020}]%
        {yang2020image}
\bibfield{author}{\bibinfo{person}{Xiaocui Yang}, \bibinfo{person}{Shi Feng},
  \bibinfo{person}{Daling Wang}, {and} \bibinfo{person}{Yifei Zhang}.}
  \bibinfo{year}{2020}\natexlab{}.
\newblock \showarticletitle{Image-text Multimodal Emotion Classification via
  Multi-view Attentional Network}.
\newblock \bibinfo{journal}{\emph{IEEE Transactions on Multimedia}}
  (\bibinfo{year}{2020}).
\newblock


\bibitem[\protect\citeauthoryear{Yap, Yolland, and Tschandl}{Yap
  et~al\mbox{.}}{2018}]%
        {yap2018multimodal}
\bibfield{author}{\bibinfo{person}{Jordan Yap}, \bibinfo{person}{William
  Yolland}, {and} \bibinfo{person}{Philipp Tschandl}.}
  \bibinfo{year}{2018}\natexlab{}.
\newblock \showarticletitle{Multimodal skin lesion classification using deep
  learning}.
\newblock \bibinfo{journal}{\emph{Experimental dermatology}}
  \bibinfo{volume}{27}, \bibinfo{number}{11} (\bibinfo{year}{2018}),
  \bibinfo{pages}{1261--1267}.
\newblock


\bibitem[\protect\citeauthoryear{Yokoya, Ghamisi, Xia, Sukhanov, Heremans,
  Tankoyeu, Bechtel, Le~Saux, Moser, and Tuia}{Yokoya et~al\mbox{.}}{2018}]%
        {yokoya2018open}
\bibfield{author}{\bibinfo{person}{Naoto Yokoya}, \bibinfo{person}{Pedram
  Ghamisi}, \bibinfo{person}{Junshi Xia}, \bibinfo{person}{Sergey Sukhanov},
  \bibinfo{person}{Roel Heremans}, \bibinfo{person}{Ivan Tankoyeu},
  \bibinfo{person}{Benjamin Bechtel}, \bibinfo{person}{Bertrand Le~Saux},
  \bibinfo{person}{Gabriele Moser}, {and} \bibinfo{person}{Devis Tuia}.}
  \bibinfo{year}{2018}\natexlab{}.
\newblock \showarticletitle{Open data for global multimodal land use
  classification: Outcome of the 2017 IEEE GRSS Data Fusion Contest}.
\newblock \bibinfo{journal}{\emph{IEEE Journal of Selected Topics in Applied
  Earth Observations and Remote Sensing}} \bibinfo{volume}{11},
  \bibinfo{number}{5} (\bibinfo{year}{2018}), \bibinfo{pages}{1363--1377}.
\newblock


\bibitem[\protect\citeauthoryear{Yu, Jiang, and Xia}{Yu et~al\mbox{.}}{2019}]%
        {yu2019entity}
\bibfield{author}{\bibinfo{person}{Jianfei Yu}, \bibinfo{person}{Jing Jiang},
  {and} \bibinfo{person}{Rui Xia}.} \bibinfo{year}{2019}\natexlab{}.
\newblock \showarticletitle{Entity-sensitive attention and fusion network for
  entity-level multimodal sentiment classification}.
\newblock \bibinfo{journal}{\emph{IEEE/ACM Transactions on Audio, Speech, and
  Language Processing}}  \bibinfo{volume}{28} (\bibinfo{year}{2019}),
  \bibinfo{pages}{429--439}.
\newblock


\bibitem[\protect\citeauthoryear{Zehtab-Salmasi, Feizi-Derakhshi,
  Nikzad-Khasmakhi, Asgari-Chenaghlu, and Nabipour}{Zehtab-Salmasi
  et~al\mbox{.}}{2021}]%
        {zehtab2021multimodal}
\bibfield{author}{\bibinfo{person}{Aidin Zehtab-Salmasi},
  \bibinfo{person}{Ali-Reza Feizi-Derakhshi}, \bibinfo{person}{Narjes
  Nikzad-Khasmakhi}, \bibinfo{person}{Meysam Asgari-Chenaghlu}, {and}
  \bibinfo{person}{Saeideh Nabipour}.} \bibinfo{year}{2021}\natexlab{}.
\newblock \showarticletitle{Multimodal price prediction}.
\newblock \bibinfo{journal}{\emph{Annals of Data Science}}
  (\bibinfo{year}{2021}), \bibinfo{pages}{1--17}.
\newblock


\bibitem[\protect\citeauthoryear{Zhai, Perez-Pozuelo, Clifton, Palotti, and
  Guan}{Zhai et~al\mbox{.}}{2020}]%
        {zhai2020making}
\bibfield{author}{\bibinfo{person}{Bing Zhai}, \bibinfo{person}{Ignacio
  Perez-Pozuelo}, \bibinfo{person}{Emma~AD Clifton}, \bibinfo{person}{Joao
  Palotti}, {and} \bibinfo{person}{Yu Guan}.} \bibinfo{year}{2020}\natexlab{}.
\newblock \showarticletitle{Making sense of sleep: Multimodal sleep stage
  classification in a large, diverse population using movement and cardiac
  sensing}.
\newblock \bibinfo{journal}{\emph{Proceedings of the ACM on Interactive,
  Mobile, Wearable and Ubiquitous Technologies}} \bibinfo{volume}{4},
  \bibinfo{number}{2} (\bibinfo{year}{2020}), \bibinfo{pages}{1--33}.
\newblock


\bibitem[\protect\citeauthoryear{Zhang, Yang, He, and Deng}{Zhang
  et~al\mbox{.}}{2020b}]%
        {zhang2020multimodal}
\bibfield{author}{\bibinfo{person}{Chao Zhang}, \bibinfo{person}{Zichao Yang},
  \bibinfo{person}{Xiaodong He}, {and} \bibinfo{person}{Li Deng}.}
  \bibinfo{year}{2020}\natexlab{b}.
\newblock \showarticletitle{Multimodal intelligence: Representation learning,
  information fusion, and applications}.
\newblock \bibinfo{journal}{\emph{IEEE Journal of Selected Topics in Signal
  Processing}} \bibinfo{volume}{14}, \bibinfo{number}{3}
  (\bibinfo{year}{2020}), \bibinfo{pages}{478--493}.
\newblock


\bibitem[\protect\citeauthoryear{Zhang, Sidib{\'e}, Morel, and
  M{\'e}riaudeau}{Zhang et~al\mbox{.}}{2020a}]%
        {zhang2020deep}
\bibfield{author}{\bibinfo{person}{Yifei Zhang},
  \bibinfo{person}{D{\'e}sir{\'e} Sidib{\'e}}, \bibinfo{person}{Olivier Morel},
  {and} \bibinfo{person}{Fabrice M{\'e}riaudeau}.}
  \bibinfo{year}{2020}\natexlab{a}.
\newblock \showarticletitle{Deep multimodal fusion for semantic image
  segmentation: A survey}.
\newblock \bibinfo{journal}{\emph{Image and Vision Computing}}
  (\bibinfo{year}{2020}), \bibinfo{pages}{104042}.
\newblock


\bibitem[\protect\citeauthoryear{Zhao, Xie, Xu, and Sun}{Zhao
  et~al\mbox{.}}{2017}]%
        {zhao2017multi}
\bibfield{author}{\bibinfo{person}{Jing Zhao}, \bibinfo{person}{Xijiong Xie},
  \bibinfo{person}{Xin Xu}, {and} \bibinfo{person}{Shiliang Sun}.}
  \bibinfo{year}{2017}\natexlab{}.
\newblock \showarticletitle{Multi-view learning overview: Recent progress and
  new challenges}.
\newblock \bibinfo{journal}{\emph{Information Fusion}}  \bibinfo{volume}{38}
  (\bibinfo{year}{2017}), \bibinfo{pages}{43--54}.
\newblock


\bibitem[\protect\citeauthoryear{Zhou, Chang, Bai, Xiao, Su, Bi, Zhang,
  Senders, Valli{\`e}res, Kavouridis, et~al\mbox{.}}{Zhou
  et~al\mbox{.}}{2019}]%
        {zhou2019machine}
\bibfield{author}{\bibinfo{person}{Hao Zhou}, \bibinfo{person}{Ken Chang},
  \bibinfo{person}{Harrison~X Bai}, \bibinfo{person}{Bo Xiao},
  \bibinfo{person}{Chang Su}, \bibinfo{person}{Wenya~Linda Bi},
  \bibinfo{person}{Paul~J Zhang}, \bibinfo{person}{Joeky~T Senders},
  \bibinfo{person}{Martin Valli{\`e}res}, \bibinfo{person}{Vasileios~K
  Kavouridis}, {et~al\mbox{.}}} \bibinfo{year}{2019}\natexlab{}.
\newblock \showarticletitle{Machine learning reveals multimodal MRI patterns
  predictive of isocitrate dehydrogenase and 1p/19q status in diffuse low-and
  high-grade gliomas}.
\newblock \bibinfo{journal}{\emph{Journal of neuro-oncology}}
  \bibinfo{volume}{142}, \bibinfo{number}{2} (\bibinfo{year}{2019}),
  \bibinfo{pages}{299--307}.
\newblock


\end{thebibliography}

\end{document}